%% file: main.tex
\documentclass[10pt,twocolumn,letterpaper]{article}

\PassOptionsToPackage{table,xcdraw}{xcolor}
\usepackage{cvpr}              


\definecolor{cvprblue}{rgb}{0.21,0.49,0.74}
\usepackage[pagebackref,breaklinks,colorlinks,allcolors=cvprblue]{hyperref}
\usepackage{graphicx}
\usepackage{amsmath}
\usepackage{amssymb}
\usepackage{booktabs}
\usepackage{tabularx}
\usepackage{array}
\usepackage{bbding}
\usepackage{enumitem}
\usepackage{multirow}
\usepackage{makecell}
\usepackage{url}
\usepackage{balance}
\usepackage[accsupp]{axessibility}

\def\paperID{10134} 
\def\confName{CVPR}
\def\confYear{2025}

\title{ImViD: Immersive Volumetric Videos for Enhanced VR Engagement }

\author{Zhengxian Yang$^{1}$\footnotemark[1], \quad  Shi Pan$^{1}$\footnotemark[1], \quad Shengqi Wang$^{1}$\footnotemark[1], \quad  Haoxiang Wang$^{1}$,\\ 
  Li Lin$^{2}$, \quad Guanjun Li$^{3}$, \quad Zhengqi Wen$^{1}$\footnotemark[2], \quad Borong Lin$^{1}$\footnotemark[2], \quad Jianhua Tao$^{1}$\footnotemark[2], \quad Tao Yu$^{1}$\footnotemark[2] \\
  $^1$Tsinghua University \;\;
  $^2$Migu Beijing Research Institute \\
  $^3$Institute of Automation, Chinese Academy of Science \\
\texttt{\small\{zx-yang23, ps23, shengqi-21, whx22\}@mails.tsinghua.edu.cn} \;\;  
\texttt{\small lilin@migu.chinamobile.com} \\
\texttt{\small guanjun.li@nlpr.ia.ac.cn}  \;\;
\texttt{\small\{zqwen, linbr, jhtao, ytrock\}@tsinghua.edu.cn}
}

\begin{document}
\twocolumn[{%
\renewcommand\twocolumn[1][]{#1}%
\maketitle
\input{sec/0_teaser}
\bigbreak
}]
\input{sec/0_abstract}  
\vspace{-10pt}
\input{sec/1_intro}
\input{sec/2_RelatedWork}
\input{sec/3_Dy-SOFT}
\input{sec/4_Benchmarks}
\input{sec/5_Results}
\input{sec/6_conclusion}
\clearpage
\section*{Acknowledgment}
This work was supported by National Key R\&D Program of China (No.2022YFF0902204), NSFC (No.62422110, 52425801), Tsinghua University-Migu Xinkong Culture Technology (Xiamen) Co.Ltd. Joint Research Center for Intelligent Light Field and Interaction Technology, Guoqiang Institute of Tsinghua University (No.2021GQG0001).
{
    \small
    \bibliographystyle{ieeenat_fullname}
    \bibliography{dynamic}
}

\input{sec/X_suppl}

\end{document}

%% file: sec/0_teaser.tex
\setlength{\tabcolsep}{0pt}
\vspace{-10pt}
\begin{tabular}{c}
\centering
\includegraphics[width=\linewidth]
{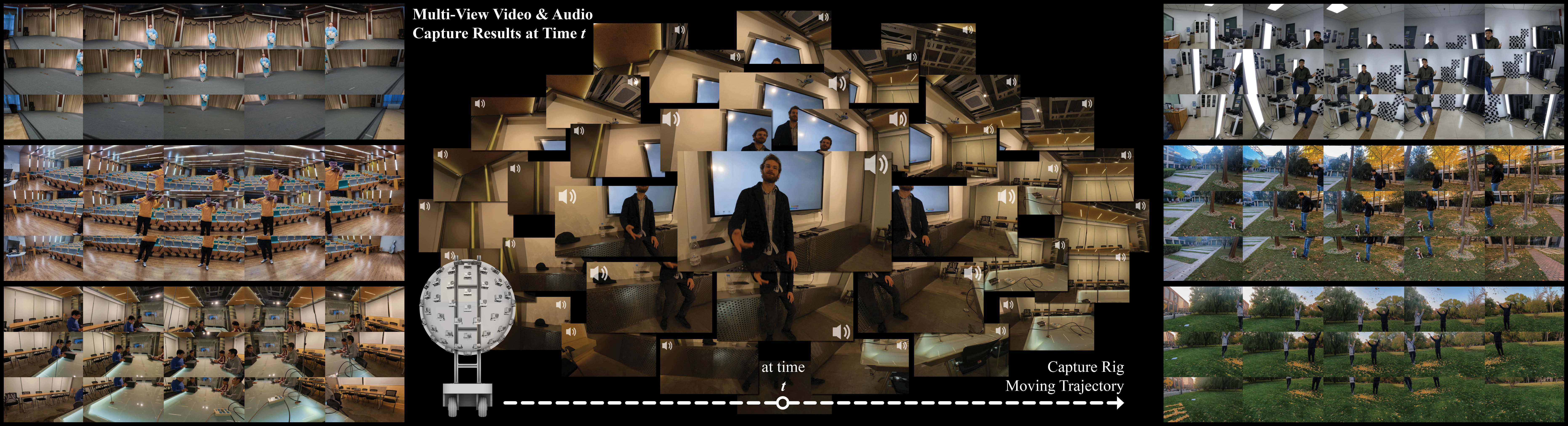} \
\end{tabular}
\captionof{figure}{
We introduce \textit{\textbf{{ImViD}}}, a dataset for immersive volumetric videos. 
\textit{ImViD} records dynamic scenes using a multi-view audio-video capture rig moving in a space-oriented manner, which provides a new benchmark for volumetric video reconstruction and its application.
}
\label{fig:teaser}

%% file: sec/0_abstract.tex
\begin{abstract}
User engagement is greatly enhanced by fully immersive multi-modal experiences that combine visual and auditory stimuli. 
Consequently, the next frontier in VR/AR technologies lies in immersive volumetric videos with complete scene capture, large 6-DoF interaction space, multi-modal feedback, and high resolution\&frame-rate contents. 
To stimulate the reconstruction of immersive volumetric videos, 
we introduce \textbf{ImViD}, a multi-view, multi-modal dataset featuring complete space-oriented data capture and various indoor/outdoor scenarios. 
Our capture rig supports multi-view video-audio capture while on the move, a capability absent in existing datasets, significantly enhancing the completeness, flexibility, and efficiency of data capture. 
\footnotetext{
\llap{\textsuperscript{*}}Equal contributions.\;  \ \llap{\textsuperscript{$\dagger$}}Corresponding author.
}
\footnotetext{Zhengxian Yang, Shengqi Wang, Haoxiang Wang, Jianhua Tao and Tao Yu are at BNRist\&Department of Automation; Shi Pan and Borong Lin are at Department of Building Science; Zhengqi Wen is at BNRist. }

The captured multi-view videos (with synchronized audios) are in 5K resolution at 60FPS, lasting from 1-5 minutes, and include rich foreground-background elements, and complex dynamics. 
We benchmark existing methods using our dataset and establish a base pipeline for constructing immersive volumetric videos from multi-view audiovisual inputs for 6-DoF multi-modal immersive VR experiences. 
The benchmark and the reconstruction and interaction results demonstrate the effectiveness of our dataset and baseline method, which we believe will stimulate future research on immersive volumetric video production.
\end{abstract}

%% file: sec/1_intro.tex
\section{Introduction}
\label{sec:intro}

High-quality and high-fidelity digital modeling of the real world has long been a topic of significant interest. With breakthroughs in 3D reconstruction algorithms and virtual reality technology, the generation of such visually compelling content is gradually becoming feasible, whether for reproducing past events and memories or remotely viewing current occurrences. Achieving complete user immersion is a goal pursued by many researchers.

    \textit{What defines an immersive volumetric video?}

Dwelling on previous tests~\cite{broxton2020immersive,apple2024} and user feedback~\cite{narciso2019immersive,gracia}, we believe that immersive media should have the following four characteristics: i) Full 360° foreground and background, ii) High-quality 6-DoF interaction, which current volumetric video can achieve, iii) Multimodal experience with light and sound, and iv) High frame rate and long-duration dynamic content. We refer to a media format that meets all these criteria as immersive volumetric video.

However, with the development of camera equipment, despite many efforts using monocular handheld cameras or fixed camera arrays to build datasets, none of these solutions fully address all four requirements. For example, Panoptic Sports~\cite{Joo_2017_TPAMI}, ZJU-Mocap~\cite{peng2021neural} and DiVa-360~\cite{lu2024diva} achieved human/object-centric 360° capture but lack complex backgrounds and multimodal data; Google’s immersive light field~\cite{broxton2020immersive} achieves 6-DoF volumetric video in an inside-looking-out manner but lacks multimodality and high frame rate, long-duration dynamic content; Replay~\cite{shapovalov2023replay} introduces sound, but the camera setup is not suitable for human viewing habits thus cannot achieve high-quality 6-DoF interaction. Both academia and industry require an appropriate capture method to support the development of immersive volumetric video technology.

Additionally, NeRF~\cite{mildenhall2021nerf}'s emergence has increased attention to reconstruction and rendering techniques, making it easier to generate high-fidelity interactive 3D scenes from multiple views. 3DGS~\cite{kerbl3DGaussianSplatting2023b} and its subsequent improvements~\cite{lu2023scaffold,yan2024street,yang2024spec,malarzgaussian,zhang2024gaussian,cheng2024gaussianpro,li2024geogaussian} have further boosted rendering speed and quality, fueling peoples' expectations for rapid real-world modeling and its applications in industries such as education, healthcare, and entertainment. This can be seen as an evolutionary opportunity for immersive volumetric video.
However, current technologies face challenges. In dynamic light field reconstruction, neither explicit~\cite{wu20244d,li2024spacetime,katsumata2024compactdynamic3dgaussian,kratimenos2023dynmf,lin2024gaussian} nor implicit neural representation based~\cite{li2022neural,song2023nerfplayer,wang2023mixed,attal2023hyperreel,fridovich2023k,cao2023hexplane,shao2023tensor4d} methods have been able to achieve a practical balance between foreground-background reconstruction accuracy, rendering speed, spatiotemporal consistency, and storage efficiency in dynamic scenes; In the realm of sound field reconstruction, it is typically necessary to collect a large amount of binaural audio or room impulse response data from different locations for model training, which is user-unfriendly in practical use.

To better support the development of immersive volumetric video technology, we have developed a moving capture rig (with 46 synchronized cameras, as shown in Fig.\ref{fig:data capture}) and designed two space-oriented data capture strategies. 
We have constructed an audiovisual dataset containing 7 indoor and outdoor scenes, with a 360° field of view, large 6-DoF interaction space, and high-resolution, long-duration dynamic contents. 
We also validate this dataset using the latest novel view synthesis and 3D reconstruction methods. Finally, a complete pipeline for constructing light and sound fields from this type of data was proposed, which initially achieved immersive volume video.

Our main contributions can be summarized as follows:

\begin{itemize}[leftmargin=*]
\item We propose to our knowledge the first and probably the highest-quality multi-view dataset \textbf{ImViD} for immersive volumetric video. It features 360° foreground and background capture, long-duration video\&audio data under high resolution@5K, high frame rate@60FPS.
\item We conduct extensive experiments to validate the effectiveness of the proposed datasets by evaluating and analyzing the performance of several current baselines. 
\item Follow the proposed pipeline shown in Figure~\ref{fig:pipeline}, We finally achieve the reconstruction of immersive volumetric video and integrate it for immersive VR experience.
\end{itemize}

%% file: sec/2_RelatedWork.tex
\section{Related Work}
\label{sec:Related Work}

\begin{table*}[h]
    \centering
    \caption{Existing real-world datasets for dynamic novel view synthesis (full version in the supplementary materials). }
    \label{tab:datasets comparison}
    \resizebox{\textwidth}{!}{%
    \renewcommand*{\arraystretch}{2}
    \begin{tabular}{m{3.5cm}cccccccccm{8cm}}
    \toprule
    \textbf{Datasets}                   & \textbf{No.Scene} & \textbf{Outdoor/Indoor} & \textbf{Cameras}                       & \textbf{Mobility}          & \textbf{Resolution} & \textbf{Angles}                          & \textbf{Duration}  & \textbf{FPS} & \textbf{Multimodality} & \textbf{Content}                                                                                                                                                                                                                       \\
    \midrule
    PanopticSports~\cite{Joo_2017_TPAMI}                      & 65                & Indoor                  & 480 cameras                            & Static                    & 640×480             & 360°                                     & 5mins              & 25           & \XSolidBrush                    & Human-centric actions                                                                                                                                                                                                                  \\
    Technicolor~\cite{sabater2017dataset}                         & 12                & Indoor                  & 16 cameras                             & Static                    & 2048×1088           & Frontal                                  & 2s                 & 30           & \XSolidBrush                     & Has a number of close-ups sequences, captured medium angle scenes and other animated scenes where the movement does not come from a human                                                                                           \\
    Immersive-Lightfield~\cite{broxton2020immersive}                & 15                & both                    & 46 cameras                             & Static                    & 2560×1920           & Frontal                                  & 10-30s                & 30           & \XSolidBrush                     & Simple and slow motion of human,animals,objects                                                                                                                                                                                        \\
    Dynamic Scene Datasets (NVIDIA)~\cite{yoon2020novel}      & 8                 & Outdoor                 & \makecell[c]{1 Mobile phone\\/12 cameras}        & \makecell[c]{Fixed-point Waving\\/Static} & 1920×1080           & Frontal                                  & 5s                 & 60           & \XSolidBrush                     & Simple body motions (facial, jump, etc)                                                                                                                                                                                                 \\
    UCSD Dynamic~\cite{lin2021deep}          & 96                & Outdoor                 & 10 cameras                             & Static                    & 1920×1080           & Frontal                                  & 1-2mins            & \textbf{120} & \XSolidBrush                     & Various visual effects and human interactions                                                                                                                                                                                          \\
    ZJU-Mocap~\cite{peng2021neural}                           & 10                & Indoor                  & 21 cameras                             & Static                    & 1024×1024           & 360°                                     & 20s                & 50           & \XSolidBrush                     & Simple body motions (punch, kick, etc.)                                                                                                                                                                                                \\
    Plenoptic Dataset (DyNeRF/Neural 3D)~\cite{li2022neural} & 6                 & Indoor                  & 21 cameras                             & Static                    & 2704×2028           & Frontal                                  & 10-30s                & 30           & \XSolidBrush                     & Contains high specularity, translucency and transparency objects, motions with changing topology, selfcast moving shadows, various lighting conditions and multiple people moving around in open living room space \\
    iPhone Datasets~\cite{gao2022monocular}                     & 14                & both                    & \makecell[c]{1 hand-held phone\\/2 cameras }          & \makecell[c]{Fixed-point Waving\\/Static} & 640×480             & Frontal                                  & 8-15s              & 30/60        & \XSolidBrush                     & Featuring non-repetitive motion, from various categories such as generic objects, humans, and pets                                                                                                                                    \\
    ENeRF-Outdoor~\cite{lin2022efficient}                       & 4                 & Outdoor                 & 18 cameras                             & Static                     & 1920×1080           & Frontal                                  & 20-40s             & 30           & \XSolidBrush                     & Complex human motions                                                                                                                                                                                                                  \\
    Replay~\cite{shapovalov2023replay}                              & 46                & Indoor                  & 12 cameras                             & Static                    & 3840×2160           & 360°                                     & 5mins              & 30           & \checkmark (Audio)             & Dancing, chatting, playing video games, unwrapping presents, playing ping pong                                                                                                                                                         \\
    360+X~\cite{chen2024360+}                               & 28                & both                    & \makecell[c]{1 360°cameras and \\1 Spectacles cameras} & Static                    & 5760×2880           & 360°                                     & 10s (2152 sequence) & 30           & \checkmark (Audio)             & Capture in 17 cities across 5 countries.Panoptic perspective to scene understanding with audio                                                                                                                                        \\
    \textbf{Im-ViD(Ours)}                       & 7                 & \textbf{both}           & \textbf{46 cameras}                    & \textbf{Moveable}          & \textbf{5312×2988}  & \textbf{Frontal and 360°} & \textbf{1-5mins}   & 60           & \textbf{\checkmark (Audio)}    & Seven common indoor and outdoor scenes in daily life, including opera, face-to-face communication, teaching, discussion, music performance, interaction with pets, and playing. Each scene has high-quality synchronized multi-view video and audio with a duration of more than 1 minute, and contains rich elements such as various small objects, glass, and changes in light and shadow                                               \\                  \bottomrule                                                                                                                                                                   
    \end{tabular}%
    }
\end{table*}

\paragraph{Datasets for Dynamic Novel View Synthesis.}
The datasets for dynamic scene reconstruction can be classified as monocular-based and multi-view-based.

Monocular acquisition systems are popular due to their low cost and ease of construction. 
such as HyperNeRF~\cite{park2021hypernerf}, Dynamic Scene Dataset~\cite{yoon2020novel}, and D2NeRF~\cite{wu2022d}. However, these datasets suffer from low resolution, limited capture space, and the limited durations. Although NeRF On-the-go~\cite{ren2024nerf} allows for larger capture ranges by walking while shooting, high-quality reconstructions are confined to the capture path, and the small field of view (FOV) limits prolonged observations of specific scene positions.

Multi-camera data collection has gained significant attention due to its ability to provide a larger FOV and richer details, such as Immersive Light Field dataset~\cite{broxton2020immersive}, Technicolor~\cite{sabater2017dataset}, UCSD Dynamic Scene Dataset~\cite{lin2021deep} and The Plenoptic Dataset~\cite{li2022neural}, followed by ~\cite{li2022streaming,lin2022efficient,wang2024masked}.
However, all these setups remain static during capture, limiting them to frontal views and hindering 360° reconstruction. Additionally, the video sequences are typically short, with a maximum duration of 2 minutes (often less than 30 seconds) and a maximum resolution of 4K, which is insufficient for immersive VR experiences. 

Moreover, the previously mentioned datasets lack sound recordings, leading to the loss of immersion without multimodality. The Replay dataset~\cite{shapovalov2023replay} addresses this by focusing on long sequences with professional actors in familiar settings. However, aside from the head-mounted cameras that enables rotation with head, all other cameras remain static. Furthermore, the DSLR arrangement does not align with human viewing habits in VR, making them unsuitable as benchmarks for novel view synthesis tasks.
Recent work~\cite{chen2024360+} presents a dataset with a 360° camera and audio capture device but constrained by a fixed-point shooting strategy, resulting in sparse viewpoints that hinder the reconstruction of high-quality dynamic scenes. More details about datasets can be found in ~\cref{tab:datasets comparison}.


\vspace{-10pt}
\paragraph{Dynamic Light\&Sound Field Reconstruction.} 

Temporal variations in input images make reconstructing dynamic scenes more challenging than static ones. The key issue is efficiently representing 4D scenes while ensuring spatiotemporal consistency and high accuracy. Traditional methods estimate varying geometries like surfaces~\cite{collet2015high} and depth maps~\cite{kanade1997virtualized} to handle dynamics. Recently, efforts have shifted towards adapting neural radiance field representations for dynamic scenes. DyNeRF~\cite{li2022neural} pioneer in the dynamic reconstruction with Neural radiance field. Followed works such as ~\cite{li2022streaming,song2023nerfplayer,wang2023mixed,attal2023hyperreel,fridovich2023k,cao2023hexplane,shao2023tensor4d} make efforts to improve the high-fidelity and efficiency in the 4D neural representation. However, the common challenges these methods face are memory cost and the modeling of spatiotemporal complexity. The latest work have extended 3DGS to dynamic scenes for high-quality real-time rendering. While modeling dynamic scenes as static ones is straightforward, methods relying solely on monocular data~\cite{jung2023deformable,liang2023gaufre} is not suitable for this paradigm, and face challenges of adding constraints or priors to better supervise training. Other works support multi-view data as input.~\cite{luiten2023dynamic} and ~\cite{sun20243dgstream} initialize each frame from the previous one to build Gaussian point clouds per-frame, leading to high memory and suddenly appearing or disappearing artifacts. 4DGS~\cite{wu20244d} introduces Hexplane to encode spatiotemporal information, using MLPs to predict Gaussian properties change. Subsequent works~\cite{duisterhof2023md,guo2024motion,li2024st} further add geometric and optical flow constraints to improve frame-to-frame motion but perform well only on short sequences with small actions. Other approaches~\cite{li2024spacetime,katsumata2024compactdynamic3dgaussian,kratimenos2023dynmf,lin2024gaussian} employ the explicit parametric form (such as radial basis functions) of 3DGS to enhance spatiotemporal consistency. Some methods~\cite{yang2023real,duan20244d} focus on representation with dual quaternions, 4D spherical harmonics, or rotors for end-to-end training. More improvements such as parallel training ~\cite{shaw2024swingsslidingwindowsdynamic} and using codebooks and masks to enhance scene compactness ~\cite{lee2024compact}, show the potential in efficient and high-quality representation of long-term complex dynamic scenes.


In addition to light field reconstruction, reconstructing the sound field from multimodal data is equally important, often referred to as novel-view acoustic synthesis. For the novel-view acoustic synthesis task, AV-NeRF ~\cite{liang2023av} synthesizes spatial audio by leveraging visual features from rendered images. NeRAF ~\cite{brunetto2024neraf} further decouples the camera and microphone poses used for training. AV-GS ~\cite{bhosale2024av} incorporates a 3D scene representation during sound field synthesis, addressing the issue where previous methods overlooked the contribution of the broader 3D scene geometry. SOAF ~\cite{gao2024soaf} considers occlusion in the light field when synthesizing the sound field. However, the above novel-view acoustic synthesis tasks typically require a large amount of multimodal training data collected from different locations, which is often user-unfriendly in practical applications. Moreover, these methods generally cannot solve the problem of sound field synthesis for moving sound sources. This paper proposes a training-free novel-view acoustic synthesis approach for our dataset, which also supports sound field synthesis for moving sound sources.

\begin{figure*}[h]
    \centering
    \includegraphics[width=0.95\textwidth]{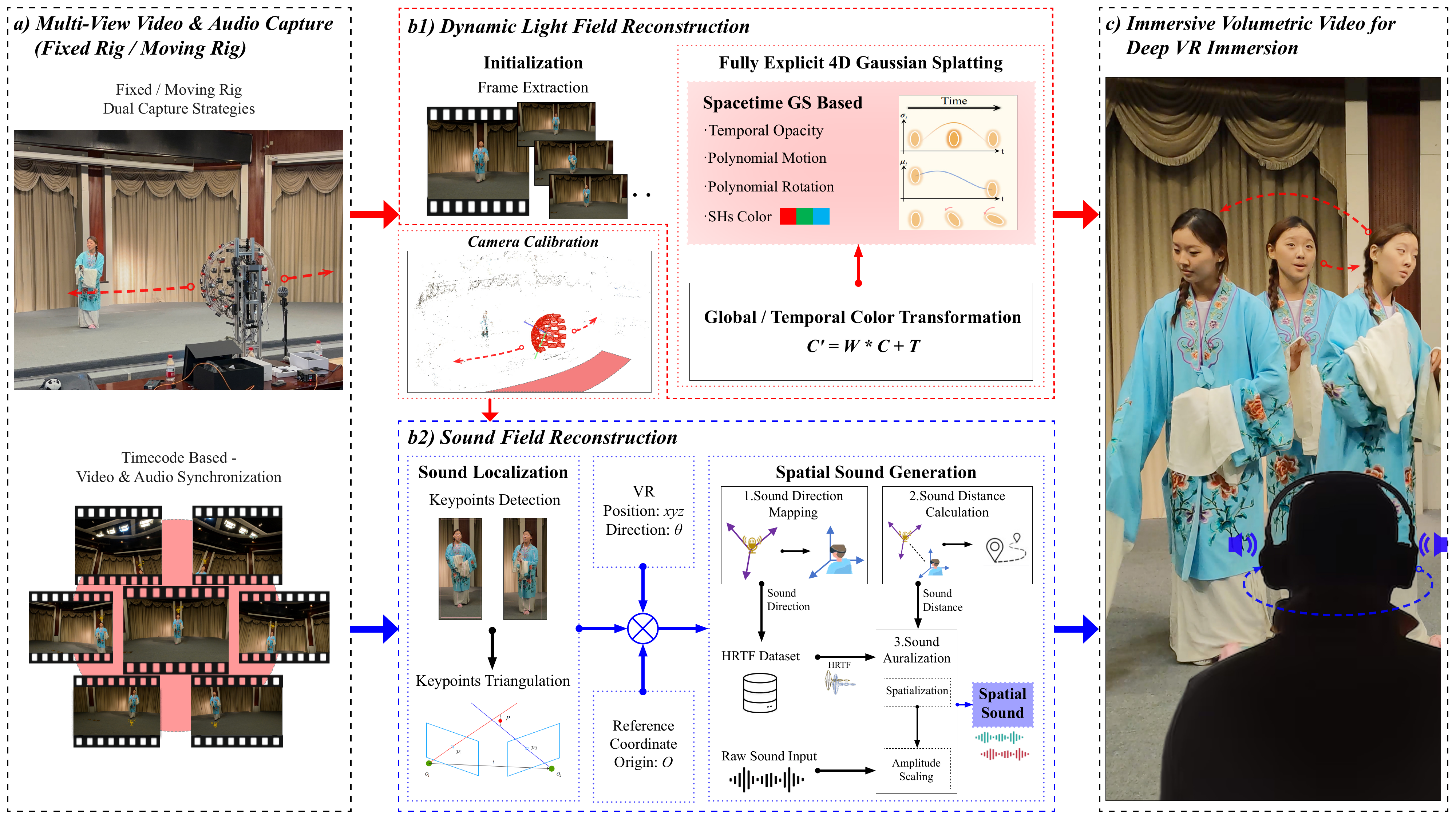}
    \caption{The pipeline to realize the multimodal 6-DoF immersive VR experiences. We applied a carefully designed rig to (a) simultaneously capture multi-view video and audio. The (b1) presents our reconstruction of dynamic light field based on STG~\cite{li2024spacetime}  while (b2) demonstrates the construction process of sound field. We have achieved better results than the original algorithm in long-term dynamic scenes by incorporating affine color transformation and t-dimensional density control. Ultimately, we achieve a 6-DoF immersive experience in both light and sound fields, and also benchmark on recent representative methods like 4DGS~\cite{wu20244d} and 4Drotor~\cite{duan20244d} to demonstrate the effectiveness of both our dataset and baseline method. }
    \label{fig:pipeline}
\end{figure*}

%% file: sec/3_Dy-SOFT.tex
\section{ImViD  Dataset}
\label{sec:dataset}

Our goal is to build an immersive volumetric video dataset of real world that integrates both foreground and background, providing assistance for the development of related research such as spatial video and VR/AR application. Therefore, we have meticulously refined every aspect of the process, from equipment setup and scene selection to shooting strategies and data processing, ensuring maintaining the elements of interest to researchers to the greatest extent.
The released ImViD dataset comprises a total of 7 large scenes, including 5 indoor and 2 outdoor environments. The content features a diverse range of common activities, such as opera, meeting, teaching, and playing on the grass. Each scene contains a set of static photos, multi-view time-synchronized video\&audio sequences at 5K resolution and 60 FPS, employing two different collection strategies, along with calibrated camera poses.

\subsection{Data Acquisition}
\begin{figure}[]
  \centering
  \includegraphics[width=\linewidth]{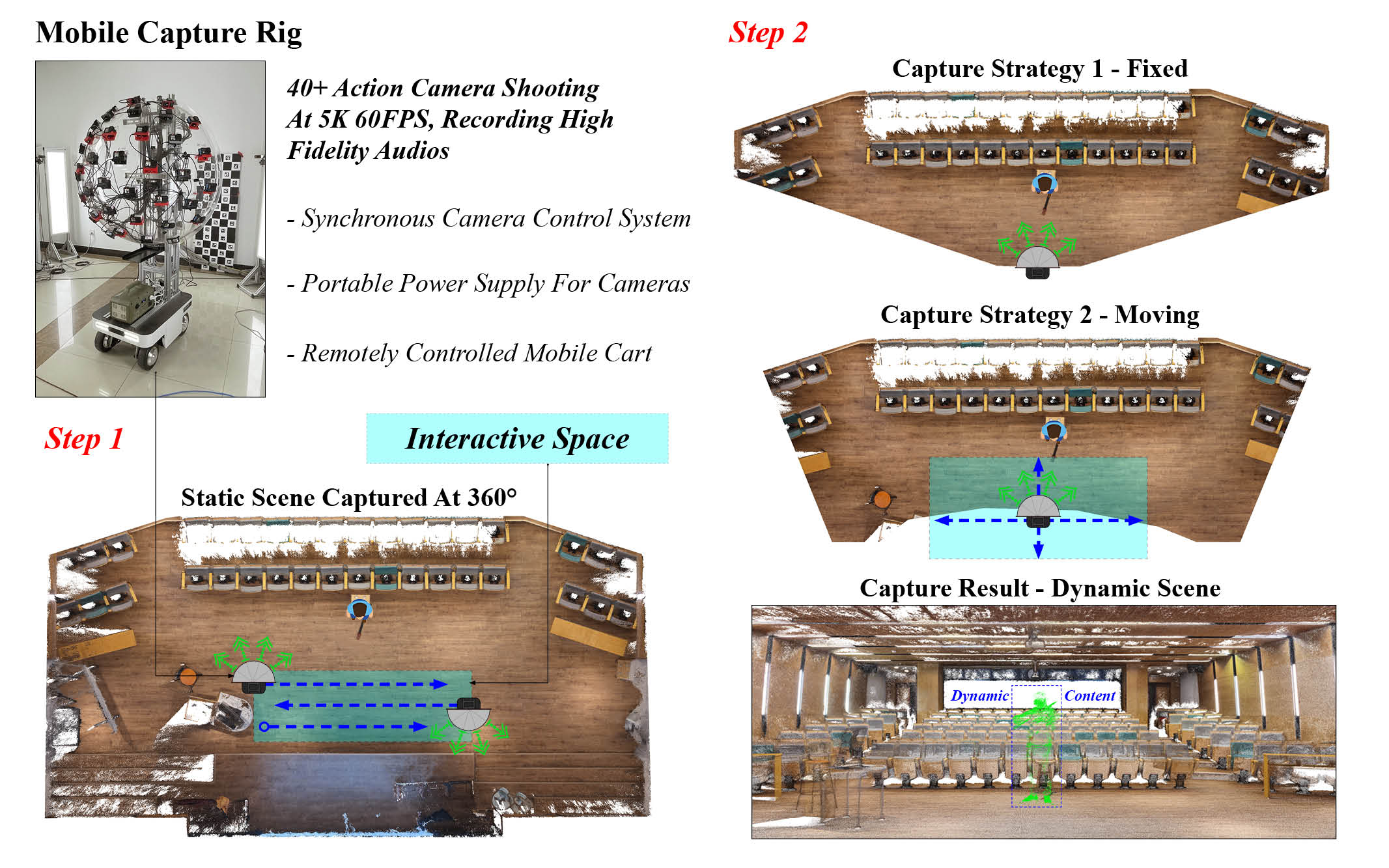}

  \caption{Our rig support two kinds of capturing strategies for high resolution, high frame rate and 360° dynamic data acquisition.}
  \label{fig:data capture}
\end{figure}

The handheld monocular camera is easy to move, providing limited perspectives from various locations. In contrast, the fixed camera arrays, while stationary, offers dense perspectives within a limited range. We aim to combine the advantages of both to design an effective data collection system and strategy for fully immersive VR experience—a feature not found in existing datasets. As shown in Figure~\ref{fig:data capture}, our capture system consists of more than 40 GoPro cameras installed on a hemispherical surface mounted on a remotely controllable mobile car, all of which are connected to a PC to receive synchronous control commands. The system is carefully designed to be at a height similar to that of a human, and the camera arrangement better aligns with human viewing habits compared to other existing devices. This rig can synchronously collect about 1,000 images at 5K resolution within a space of at least 6 m³ in 2 min. It is also capable of synchronously capturing video and audio at 60 FPS and 5K resolution for 30 min (limited by heat dissipation). Thus, our capture strategy is as follows: 

\noindent\textbf{Step1:} Efficient high-quality, high-density 360° image acquisition of static environments.

\noindent\textbf{Step2:} Continuous synchronous collection of dynamic scenes under two different strategies.

i) Fixed-point shooting. The cart remains stationary during the capture, focusing on capturing dynamic details.

ii) Moving shooting. The cart slowly moves while the scene is in motion, providing additional level of details and larger exploratory space.

Throughout the capture process, we ensure strict synchronization of the world time codes across all cameras. We carefully maintained consistent focal lengths, exposure, and white balance settings for each scene, based on the results of manual iterative testing of camera settings, thereby minimizing hardware-induced variations in data quality. To achieve high-quality audio, we increased the noise reduction coefficient of the cameras and minimized noise generation as much as possible. 

\begin{table*}[]
    \centering
    \caption{The detailed statistics of our ImViD dataset. \textbf{\#Take} represents the number of different episodes we collected in this scene. \textbf{\#Strategy} represents the strategies used in Step2. 1 indicates only (i) was used in Step 2, 2 indicates that both were used. \textbf{Avg.S-T Density} indicates the spatiotemporal coverage density of our collection method, measured by the captured volume size per second.}
    \label{tab:dataset_statistic}
    \resizebox{\linewidth}{!}{%
    \begin{tabular}{lcccccccc}
    \toprule
    \multicolumn{1}{c}{Name}           & \#Camera & \#Static viewpoints & \#Take & \#Strategy  & Avg.S-T Density(m³/s) & Viewing Space & Avg.Duration & Storage(GB) \\ \midrule
    Scene1 Opera                      & 39       & 1152                & 2      & 1                 & -                      & 180°    & 3min22s      & 226         \\
    Scene2 Laboratory                 & 39       & 1225                & 2      & 2                 & 0.10                   & 360°    & 1min42s      & 137.3       \\
    Scene3 Classroom                  & 39       & 1223                & 2      & 2                 & 0.10                   & 360°    & 4min42s      & 497         \\
    Scene4 Meeting                    & 39       & 1223                & 1      & 1                 & -                      & 360°    & 3min16s      & 114         \\
    Scene5 Rendition                  & 39       & 1620                & 4      & 2                 & 0.10                   & 360°    & 2min02s      & 516         \\
    Scene6 Puppy                        & 39       & 1404                & 3      & 2                        & 0.10                   & 360°  & 1min50s        & 359            \\
    Scene7 Playing                   & 39       & 1224                    & 2       & 2                        & 0.10                   & 360° & 1min10s         & 220            \\
    \multicolumn{1}{c}{\textbf{Total}} & -        & -                   & 16     & -                & -                      & -         & 38min46s    & 2069.3      \\ \bottomrule
    \end{tabular}%
    }
\end{table*}

\subsection{Data Processing}
For the photos from Step1, we completed the calibration in COLMAP~\cite{schoenberger2016sfm} by combining the cameras' prior parameters and their relative positions. For the video\&audio from Step2, we first aligned the multi-view sequences for each scene using the world time codes. This step is crucial as the quality of the alignment directly determines the quality of dynamic scene reconstruction and significantly impacts whether the multi-view audio can be used for constructing a 6-DoF sound field. Our time code alignment error is at the level of approximately 2 milliseconds. Subsequently, we performed frame extraction from the videos, again utilizing the GoPro camera parameters as priors to complete the calibration.

It is important to note that for the dynamic scene data captured with a moving rig, we also performed time code alignment. However, this data poses significant challenges for existing calibration methods, often resulting in errors and floaters. There are also no solutions for reconstruction algorithms specifically addressing dynamic scenes from moving rig currently. Therefore, we do not provide the calibrated results for this type of data at this time. But we believe that this data will greatly contribute to the advancement of the field, and thus we will also publicly releasing the data which has been time-synchronized.

\subsection{Dataset Analysis}
\noindent\textbf{Diversity.} 
Our dataset, as depicted in Table~\ref{tab:dataset_statistic}, includes 7 common indoor and outdoor scenes, each comprising one or more takes, totaling 16 video sequences. Each scene is rich in foreground and background elements, meticulously crafted to preserve the authenticity of the environment. The characters, clothing, and motions within each scene exhibit a diverse range, set against varying ambient lighting conditions. We also included areas of interest like tables, chairs, windows as much as possible for researchers.

\noindent\textbf{Quality.} 
Leveraging a total of 39 cameras, we captured dynamic scenes at a resolution of 5312×2988 and 60 FPS. The images of static environment is 5568×4176.

\noindent\textbf{Duration.}
As outlined in Table~\ref{tab:dataset_statistic}, each scene in our dataset contains at least 1 minute of footage. The longest sequence in our forthcoming public dataset spans approximately 5 minutes. The total processed data duration amounts to 38 minutes 46 seconds, including 139,560 frames per camera.

\noindent\textbf{Coverage.}
\begin{figure}
    \centering
    \includegraphics[width=\linewidth]{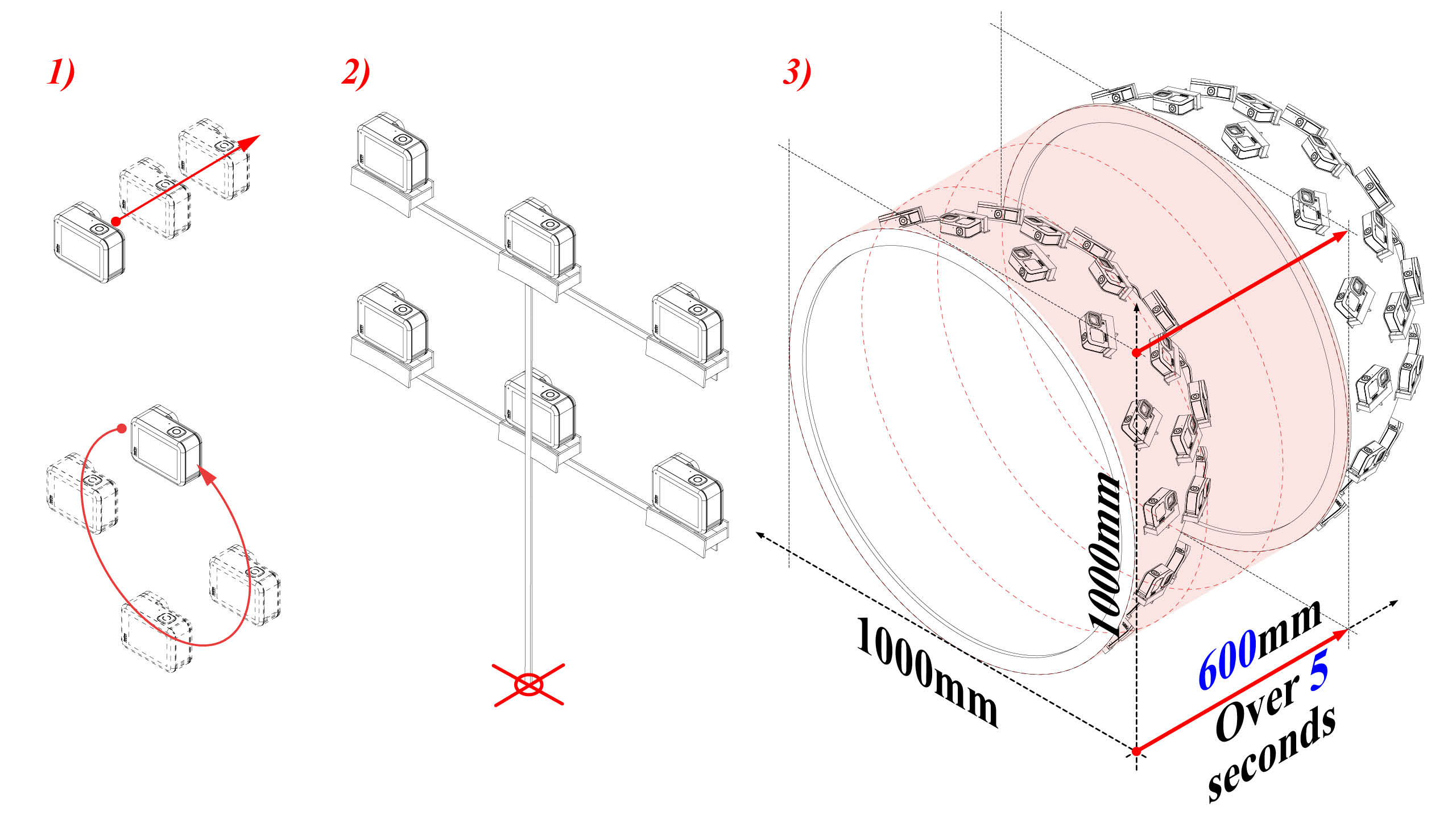}
    \caption{Calculation method for \textbf{spatiotemporal capture density}. 1) the capture strategy of handheld monocular camera 2) represents the fixed camera array 3) Our rig covers a volume of 0.6× $\pi$ ×$0.5^2$ m³ over 5 seconds.}
    \label{fig:spatialtemporaldensity}
\end{figure}
As highlighted in Table~\ref{tab:dataset_statistic}, we conducted image acquisition from over 1,000 viewpoints before recording dynamic content, thereby offering a 180°-360° viewing space in VR. For scenes employing the mobile shooting strategy, we introduced a metric termed 'spatiotemporal capture density,' quantifying the captured volume covered by our rig per second, as shown in Figure~\ref{fig:spatialtemporaldensity}. This metric, unavailable in current datasets (where static camera arrays are deemed 0 and handheld monocular cameras follow a single trajectory without capturing volume), registers at approximately 0.10 $m^3/s$ in our dataset.

%% file: sec/4_Benchmarks.tex
\section{Benchmarks and Experiments}
\label{sec:benchmarks}

To comprehensively assess the effectiveness of the dataset, we selected three scenes (2 indoor and 1 outdoor) to test the latest dynamic light field reconstruction algorithms, providing a benchmark for research in areas such as novel view synthesis. We also proposed an improvement strategy based on one of the methods. Furthermore, we presented and validated a feasible pipeline for constructing 6-DoF sound fields from multi-view audio sequences. In this section, we will first introduce the selected light field pipeline and our proposed sound field pipeline (Sec~\ref{subsec:baseline}), then provide details on the experimental setup (Sec~\ref{subsec:settings}).

\subsection{Baseline}
\label{subsec:baseline}
\paragraph{Dynamic Light Field Reconstruction.} We conducted a thorough survey of the current dynamic reconstruction algorithms, prioritizing rendering speed and support for multi-view data input. We finally decided to establish a benchmark on 3DGS-based method from three paradigms.

4DGS~\cite{wu20244d} introduces Hexplane to encode spatiotemporal information to feature $f_d$ and use MLPs decoders $D=\left \{ \phi_x,\phi_r,\phi_s \right \} $ to predict gaussians' changes in position $\Delta \chi=\phi_x(f_d)$ , rotation $\Delta r=\phi_r(f_d)$, and scaling $\Delta s=\phi_s(f_d)$. Utilizing the sparse point cloud from frame 0 as initialization and training static gaussians for the first 3000 iterations, the dynamic scene $G$ is represented as:
\begin{equation}
    G(\chi',r',s')=G(\chi+\Delta \chi ,r+\Delta r, s+\Delta s)   
\end{equation}

4DrotorGS~\cite{duan20244d} is a representative method of dimensionality expansion. It extends 3DGS to a four-dimensional Gaussian ellipsoid by introducing the concept of rotor, expanding the covariance matrix to four dimensions, i.e.:
\begin{equation}
    G_{4D}(x)=e^{-\frac{1}{2}(x-\mu_{4D})^{T}\Sigma _{4D}^{-1}(x-\mu_{4D}) }  
\end{equation}
Each frame of the scene corresponds to a time slice of the 4D Gaussian ellipsoids followed by $\alpha$-blending.

STG~\cite{li2024spacetime}, on the other hand, use polynomial fitting for the motion and rotation of the 3DGS, and employs an RBF function to model the opacity changes over time. It concatenates the sparse point clouds from all frames as the initialization, and the final representation of each point in the scene is:
\begin{equation}
    \alpha _{i}(t)=\sigma _{i}(t) \exp(-\frac{1}{2}(x-\mu_{i}(t) )^{T}\Sigma _{i}(t)^{-1}  (x-\mu_{i}(t) ))
\end{equation}

Although we ensured consistent white balance and other settings across cameras based on the characteristics of scenes during shooting, color differences still exist due to factors like occlusions when capturing from different angles. During experiments, we found that while existing methods fit well to the training views, there is a noticeable color shift during viewpoint transitions in the reconstructed light field, manifesting as flickering and floaters. This is especially apparent when training in segments, where the differences between segments are more pronounced, significantly reducing immersion. We improved the original STG (named as STG++) by introducing global, time-invariant color mapping during training. Detailed implementation can be seen in supplementary materials.

\vspace{-10pt}
\paragraph{Sound Field Reconstruction.}
Taking the recording microphone as the origin of the coordinate system, we denote the coordinates of the sound source as $s(t)=\{x_s(t),y_s(t)\}$, the user wearing the VR device as $l(t)=\{x_l(t), y_l(t)\}$, and the angle of deviation from the y-axis in the counterclockwise direction as $\theta_l(t)$, where $t$ is the time index. Based on these premises, generating the user's binaural spatial audio can be divided into three parts: sound direction mapping, sound distance mapping, and sound auralization. 
Note that we assume there is only one dominating sound source (omnidirectional emitting) to simplify the sound field reconstruction. 

In the sound direction mapping part, we calculate the direction of the sound source relative to the user's ears, $\theta_s(t)$,
\begin{equation}
    \theta_s(t) = \arccos\frac{\mathbf{v}_1(t)\mathbf{v}_2(t)}{|\mathbf{v}_1(t)||\mathbf{v}_2(t)|},
\end{equation}
where
\begin{align}
\mathbf{v}_1(t) &= [x_s(t)-x_l(t), y_s(t)-y_l(t)]^T,\\
\mathbf{v}_2(t) &= [-\sin{\left(\theta_l(t)\right)}, \cos{\left(\theta_l(t)\right)}]^T,
\end{align}
In the sound distance mapping part, we calculate the scaling of the sound source when it reaches the user's ears, relative to the original recording. We denote this scaling parameter as $\lambda(t)$,
\begin{align}
    \lambda(t) = \frac{\sqrt{x_s^2(t)+y_s^2(t)}}{\sqrt{(x_s(t)-x_l(t))^2+(y_s(t)-y_l(t))^2}}
\end{align}
In the sound auralization part, we convert the original recording into spatial audio based on $\theta_s(t)$ and $\lambda(t)$.
In the short-time Fourier transformation (STFT) domain, we denote the audio from the left and right ears as $A_L(t,f)$ and $A_R(t,f)$, respectively, where $f$ is the frequency index. Then $A_L(t,f)$ and $A_R(t,f)$ can be written as
\begin{align}
    A_L(t,f) &= F_L(\theta_s(t)) F_l(t,f)  A_O(t,f) \label{left}\\   A_R(t,f) &= F_R(\theta_s(t)) F_l(t,f) A_O(t,f), \label{right}
\end{align}
where $F_L(\theta_s(t))$ and $F_R(\theta_s(t))$ are head-related transfer functions (HRTF) based on $\theta_s(t)$, $F_l(t)$ denotes the room impulse response (RIR) at the user's location and $A_O(t,f)$ is the sound emitted from the speaker in the STFT domain.
We utilize $\theta_s(t)$ to retrieve $F_L(\theta_s(t))$ and $F_R(\theta_s(t))$ from the SADIE II dataset~\cite{armstrong2018perceptual}.
In order to model $F_l(t)$ and $A_O(t,f)$, we make the following assumptions: $F_l(t)$ change between different location at the same time index $t$ only depending on the parameter $\lambda(t)$, i.e.,
\begin{align}
    F_l(t,f) = \lambda(t) F(t,f),
\end{align}
where $F(t)$ is the RIR at the recording microphone.
Using the above assumption, Eq.(\ref{left}) and Eq.(\ref{right}) can be rewritten as 
\begin{align}
        A_L(t,f) &= \lambda(t) F_L(\theta_s(t)) A(t,f), \\   
        A_R(t,f) &= \lambda(t) F_R(\theta_s(t)) A(t,f),
\end{align}
where $A(t,f)=F(t,f) A_O(t,f)$ is the audio at the recording microphone.

\subsection{Experimental Settings}
\label{subsec:settings}
\noindent\textbf{Training setups.} The training for all light field data was completed on a single NVIDIA A100 GPU. Due to memory limitations, we split the video and trained a model every 60 frames, total 300 frames. We also followed the original train-test-split strategy of the baseline for our experiments.

\noindent\textbf{Metric.} We analyzed the quality of dynamic light field reconstruction using the most common evaluation metrics \textbf{PSNR}, \textbf{SSIM}, and \textbf{LPIPS(alex)}.

%% file: sec/5_Results.tex
\section{Results and Analysis}
\label{sec:Results}

\subsection{Validation of Light Field Data}
\begin{figure*}[h]
    \centering
    \includegraphics[width=\textwidth]{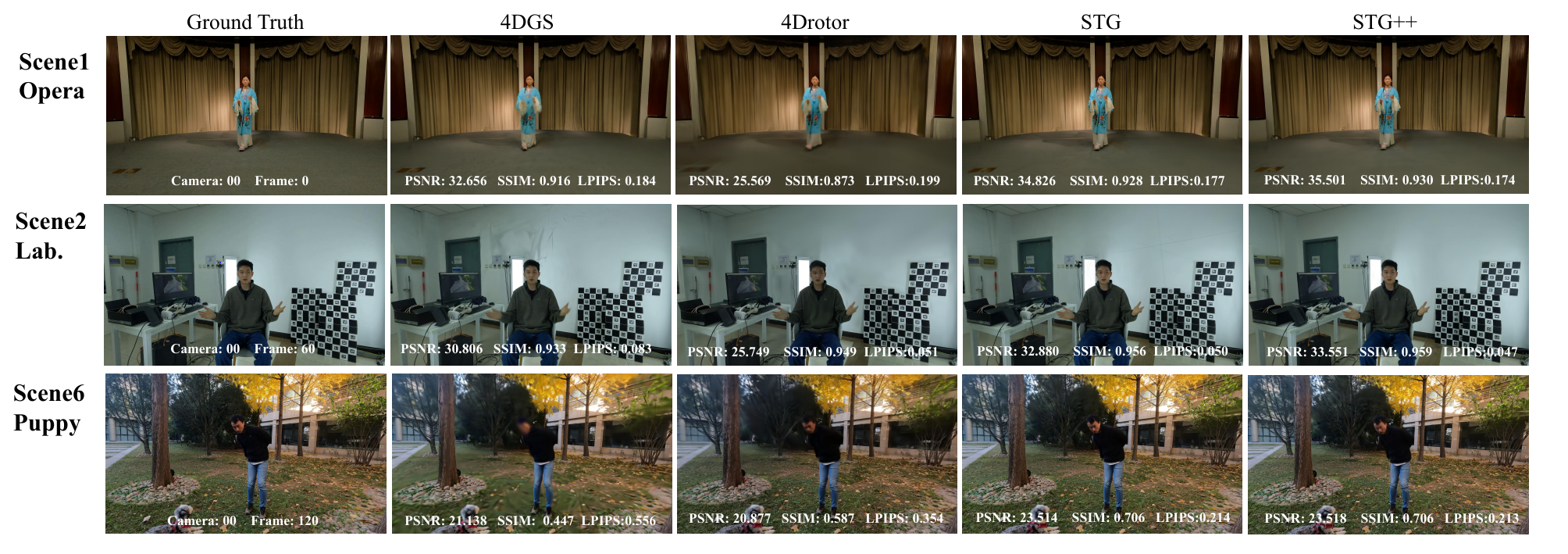}
    \caption{Comparison of the rendering results of four baselines on \textit{Scene 1 Opera}, \textit{Scene 2 Laboratory}, and \textit{Scene 6 Puppy}. }
    \label{fig:benchmark}
\end{figure*}

\paragraph{Quantitative Results.}
As illustrated in Table~\ref{tab:benchmark}, \ref{tab:memory cost} and Figure~\ref{fig:benchmark}, our indoor and outdoor data both yield correct results across three 3DGS-based paradigms. STG++ outperforms all baselines in all scenes, delivering slightly better results, the highest rendering speed, and a moderate model size. We also found that, since 4DGS trains using the sparse point cloud from the 0th frame and limits the number of final points, its performance is significantly lower than other pipelines on texture-richer outdoor scenes, such as \textit{Scene6 puppy}. Additionally, due to the density control in the time dimension implemented by 4Drotor, it performs better in frames and regions with more significant motion, which is an aspect worth referencing and adopting in future work. More results can be seen in the supplementary material.
\begin{table}[]
    \centering
    \caption{Test views performance of 3DGS-based dynamic scene reconstruction method on ImViD dataset. 
    }
    \label{tab:benchmark}
    \resizebox{\columnwidth}{!}{%
    \begin{tabular}{lccclllccclll}
    \toprule
    \multicolumn{1}{c}{\multirow{2}{*}{Method}} & \multicolumn{3}{c}{Scene1 Opera\_girl} & \multicolumn{3}{c}{Scene2 Laboratory}     & \multicolumn{3}{c}{Scene6 Puppy}                                                    \\ \cline{2-10} 
    \multicolumn{1}{c}{}                        & PSNR↑        & SSIM↑      & LPIPS↓      & \multicolumn{1}{c}{PSNR↑} & \multicolumn{1}{c}{SSIM↑} & \multicolumn{1}{c}{LPIPS↓} & PSNR↑                & SSIM↑                & LPIPS↓               \\ \midrule
    4DGS                                        & 23.227       & 0.753      & 0.410       & 25.798                    &  0.890                    & 0.176                      & 18.121               & 0.222                & 0.711                \\
    4DRotor                                     & 27.263       & 0.775      & 0.328       & 28.007                    &  0.917                    & 0.098                      & 17.916               & 0.298                & 0.331                 \\
    STG                                         & 28.482       & 0.786      & 0.287       & 26.306                    &  0.910                    & 0.114                      & 20.497               & 0.594                & 0.211                \\
    STG++                                       & 31.240       & 0.799      & 0.277       & 27.581                    &  0.916                    & 0.107                      & 20.533               & 0.597                & 0.202                \\ \bottomrule
    \end{tabular}%
    }
\end{table}

\begin{figure}
    \centering
    \includegraphics[width=\linewidth]{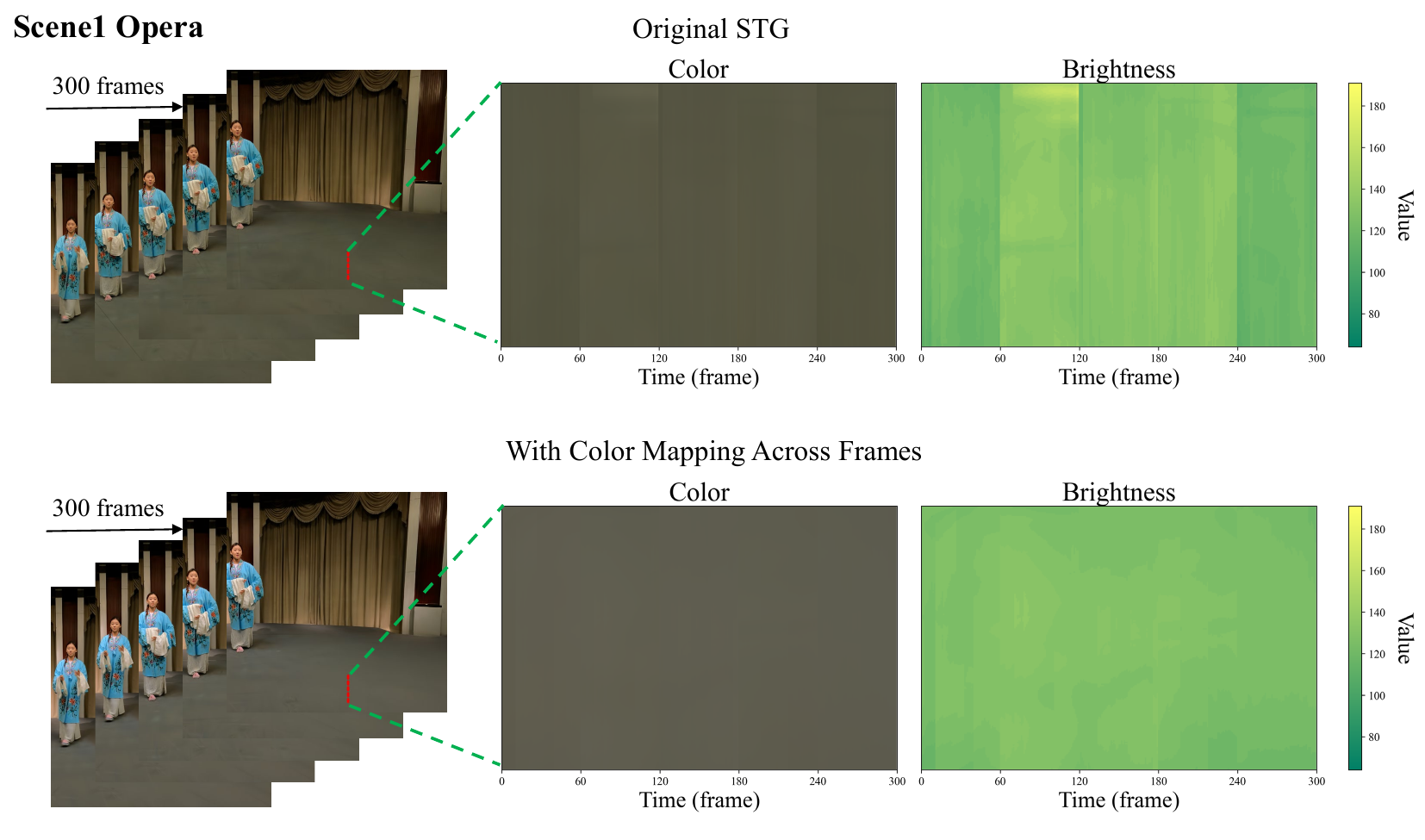}
    \caption{The continuity of pixels at the same location across different frames and segments. The visualizations present the RGB (middle) and brightness (right) variations.}
    \label{fig:color_mapping}
\end{figure}

\vspace{-10pt}
\paragraph{Qualitative Results.}
As shown in Figure~\ref{fig:color_mapping}, Our STG++ achieves better temporal continuity in the color of pixels at the same location. There are almost no abrupt transitions both within and between segments.  Details are showed in the supplementary for a more intuitive experience.

\begin{table}[]
    \centering
    \caption{The storage size of training results from different methods and their rendering speed on a single A100 GPU. *CPU version.
    }
    \label{tab:memory cost}
    \resizebox{\columnwidth}{!}{%
    \begin{tabular}{lcccccc}
    \toprule
                 & \multicolumn{2}{c}{Scene1 Opera\_girl} & \multicolumn{2}{c}{Scene2 Laboratory}              & \multicolumn{2}{c}{Scene6 Puppy}                   \\ \cline{2-7} 
                 & Mem(MB)            & FPS                & Mem(MB)                 & FPS                     & Mem(MB)                 & FPS                       \\ \midrule
    4DGS         & 101.05             & 46.22              & 105.02                  & 44.93                   &  101.65                 & 41.86                           \\
    4DRotor*     & 5818.61            & 1.00               & 6867.18                 & 0.67                    &  4415.82                & 0.56                        \\
    STG          & 382.41             & 96.10              & 318.41                  & 133.65                  &  1287.55                & 100.00                          \\
    STG++        & 387.17             & 108.89             & 329.07                  & 137.78                  &  1276.21                & 110.47                         \\ \bottomrule
    \end{tabular}%
    }
\end{table}

\subsection{Effectiveness of Sound Field Data}

\begin{table}[]
    \centering
    \caption{User study for the sound field construction. 
    }
    \label{tab:sound_res}
    \resizebox{\linewidth}{!}{%
    \begin{tabular}{lccc}
    \toprule
          &{Auditory Spatial Perception}   & {Sound Quality}      & Immersiveness \\ \midrule
Very poor & 0.00\%   & 0.00\%   & 0.00\%  \\
Poor      & 0.00\%   & 0.00\%    & 0.00\%  \\
Fair      & 14.28\%  & 19.04\%   & 9.52\%   \\
Good      & 23.80\%  & 33.33\%  & 42.85\%   \\
Excellent & 61.90\%  & 47.61\%   & 47.61\%  \\ \bottomrule
    \end{tabular}%
    }
\end{table}

\paragraph{Quantitative Results.}
Since our sound field reconstruction algorithm is non-trainable, the reconstructed spatial audio lacks a corresponding ground truth, making it difficult to evaluate the results using quantitative metrics. Therefore, we conducted a user study with 21 experts to assess the performance of the sound field reconstruction. 

Each participant rated the sound field reconstruction on three dimensions: auditory spatial perception, sound quality, and immersiveness. Auditory spatial perception refers to the listener's ability to perceive the distribution and localization of sound in space. Sound quality refers to the audio quality of the spatial audio compared to the corresponding microphone-recorded signal. Immersiveness refers to the listener's sense of being in a space where the sound source genuinely exists.

The results of the user study are shown in Table~\ref{tab:sound_res}, with numbers as percentages of participants. 
The majority of the participants (61.90\%) rated the auditory spatial perception as excellent, 80.94\% participants felt that the generated spatial audio did not exhibit significant degradation in quality, and 90.46\% found the audio immersive, which demonstrates the effectiveness of our data capture and sound field reconstruction methods.


\subsection{Multimodal VR Experiences.}
\begin{figure}
    \centering
    \includegraphics[width=0.93\linewidth]{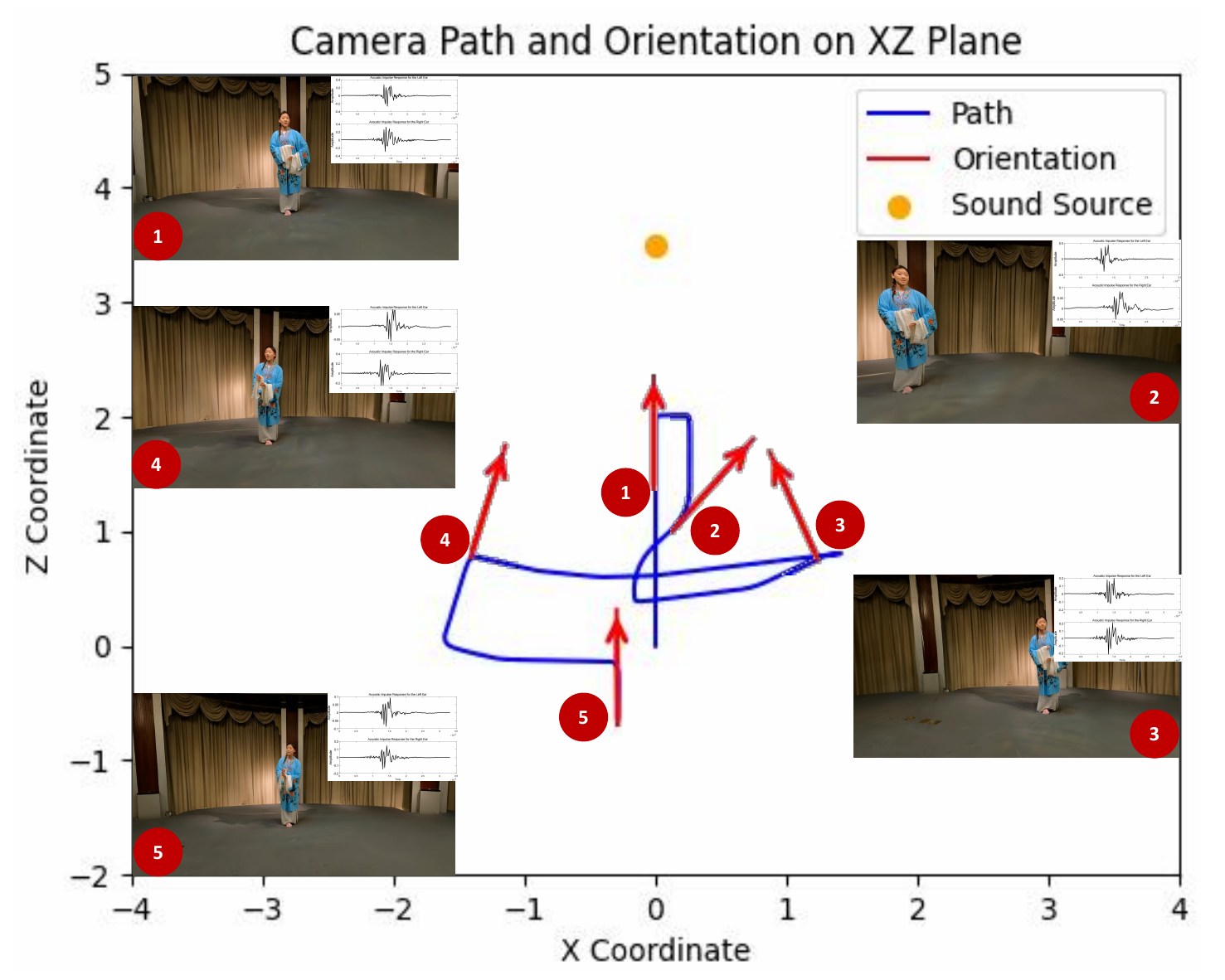}
    \caption{Visualization of the interaction trajectory and corresponding visual\&auditory results. Note that the multi-modal feedback changes consistently with the positions and orientations.}
    \label{fig:application}
\end{figure}

We finally integrated light field and sound field reconstruction results to achieve a 6-DoF multimodal immersive VR experience with real-time rendering speed at 60 FPS on a single 3090 GPU. As shown in Figure~\ref{fig:application} , the visual and auditory experiences correspond uniquely to the user's position and viewing direction. The experience can be further explored by listening through headphones with the supplementary video provided.

%% file: sec/6_conclusion.tex
\section{Conclusion}
\label{sec:conclusion}

In this work, we introduce ImViD, to our knowledge, the first dataset for immersive volumetric video. 
This dataset encompasses seven large indoor and outdoor scenes, featuring rich foreground and background content that showcases engaging and commonly encountered daily scenarios, along with high-quality synchronized audio sequences. 
We validated the effectiveness of the dataset using the latest dynamic light field reconstruction algorithms, providing both quantitative and qualitative analyses of current limitations and potential improvements. Moreover, we propose a base pipeline for constructing immersive volumetric video from this data. 
We believe ImViD will facilitate the exploration of the limitations of current volumetric video reconstruction algorithms and drive immersive VR/AR experiences.

%% file: sec/X_suppl.tex
\clearpage
\setcounter{page}{1}
\maketitlesupplementary

\appendix
\section{Overview}
Within the supplementary material, we provide:
\begin{itemize}
\item A more detailed introduction and analysis of existing datasets for Dynamic Novel View Synthesis (NVS) tasks in Appendix~\ref{appendixB};
\item More benchmark results and discussion in Appendix~\ref{appendixC}; 
\item Additional experiments details and STG++ implement details in Appendix~\ref{appendixD}.
\item Some clarifications and more descriptions of technical details regarding capture rig in Appendix~\ref{appendixE}.
\item Real-time immersive volumetric video demos and other dynamic scene reconstruction results are in our video. You can see a video demo on our homepage.
\end{itemize}

\section{Comprehensive Summary of Datasets for Dynamic Novel View Synthesis Tasks}
\label{appendixB}

The earliest studies on dynamic reconstruction have naturally focused on human digital avatars. Datasets such as Human3.6M~\cite{ionescu2013human3}, Panoptic Sports~\cite{Joo_2017_TPAMI}, ZJ-Mocap~\cite{peng2021neural}, and Tensor-4D~\cite{shao2023tensor4d} primarily focus on depicting simple human actions but do not include backgrounds, which is crucial to the immersive application experiences. We will introduce more complex datasets that include environments from monocular based and multi-view based.

Monocular acquisition systems are popular due to their low cost and ease of construction. Datasets such as HyperNeRF~\cite{park2021hypernerf}, Dynamic Scene Dataset~\cite{yoon2020novel}, and D2NeRF~\cite{wu2022d} use a mobile phone as devices, capturing dynamic scenes by waving the phone. However, these datasets suffer from resolutions below 1080p, limited capture space (similar to fixed-point shooting), and durations under one minute. Although NeRF On-the-go~\cite{ren2024nerf} allows for larger capture ranges by walking while shooting, high-quality reconstructions are confined to the vicinity of the capture path, and the small field of view (FOV) limits prolonged observations of specific scene positions.

Multi-camera data collection has gained significant attention due to its ability to provide a larger FOV and richer details. For instance, the Immersive Light Field dataset~\cite{broxton2020immersive} employs 46 cameras to capture 15 indoor and outdoor scenes, while Technicolor~\cite{sabater2017dataset} uses a 4×4 camera rig for 12 indoor sequences. The UCSD Dynamic Scene Dataset~\cite{lin2021deep} consists of 96 outdoor videos focused on single-person activities captured by 10 cameras. The Plenoptic Dataset~\cite{li2022neural} uses 21 cameras for 6 indoor scenes. Similarly, datasets like ~\cite{li2022streaming,lin2022efficient,wang2024masked} utilize 13, 18, and 24 cameras, respectively, to capture dynamic scenes. However, all these setups remain static during capture, limiting them to frontal views and hindering 360° reconstruction. Additionally, the video sequences are typically short, with a maximum duration of 2 minutes (often less than 30 seconds) and a maximum resolution of 3840×2160, which is insufficient for immersive VR experiences. 

Moreover, the previously mentioned datasets, whether monocular or multi-view, lack sound recordings, despite the importance of multimodality for immersion. The Replay dataset~\cite{shapovalov2023replay} addresses this by focusing on long sequences with professional actors in familiar settings. It employs a ring of 8 static DSLR cameras paired with binaural microphones and 3 head-mounted GoPro cameras, providing 46 videos at 4K. However, aside from the head-mounted cameras, which can rotate slightly with head movements, all other cameras remain static. Furthermore, the DSLR arrangement does not align with human viewing habits in VR, making them unsuitable as benchmarks for novel view synthesis tasks.
The latest work~\cite{chen2024360+} presents a dataset of 28 scenes captured with a 360° camera, each including multiple audio and video sequences. However, this dataset is constrained by a fixed-point shooting strategy, resulting in sparse viewpoints that hinder the reconstruction of high-quality dynamic scenes. Further comparisons between our work and these datasets can be found in ~\cref{tab:full datasets comparison}.

\section{More Benchmark Results and Analysis}
\label{appendixC}
For fair evaluation, all of our experiments use the default parameters recommended by these works. 
\subsection{Quantitative \& Qualitative Results}
\begin{figure*}
    \centering
    \includegraphics[width=\textwidth]{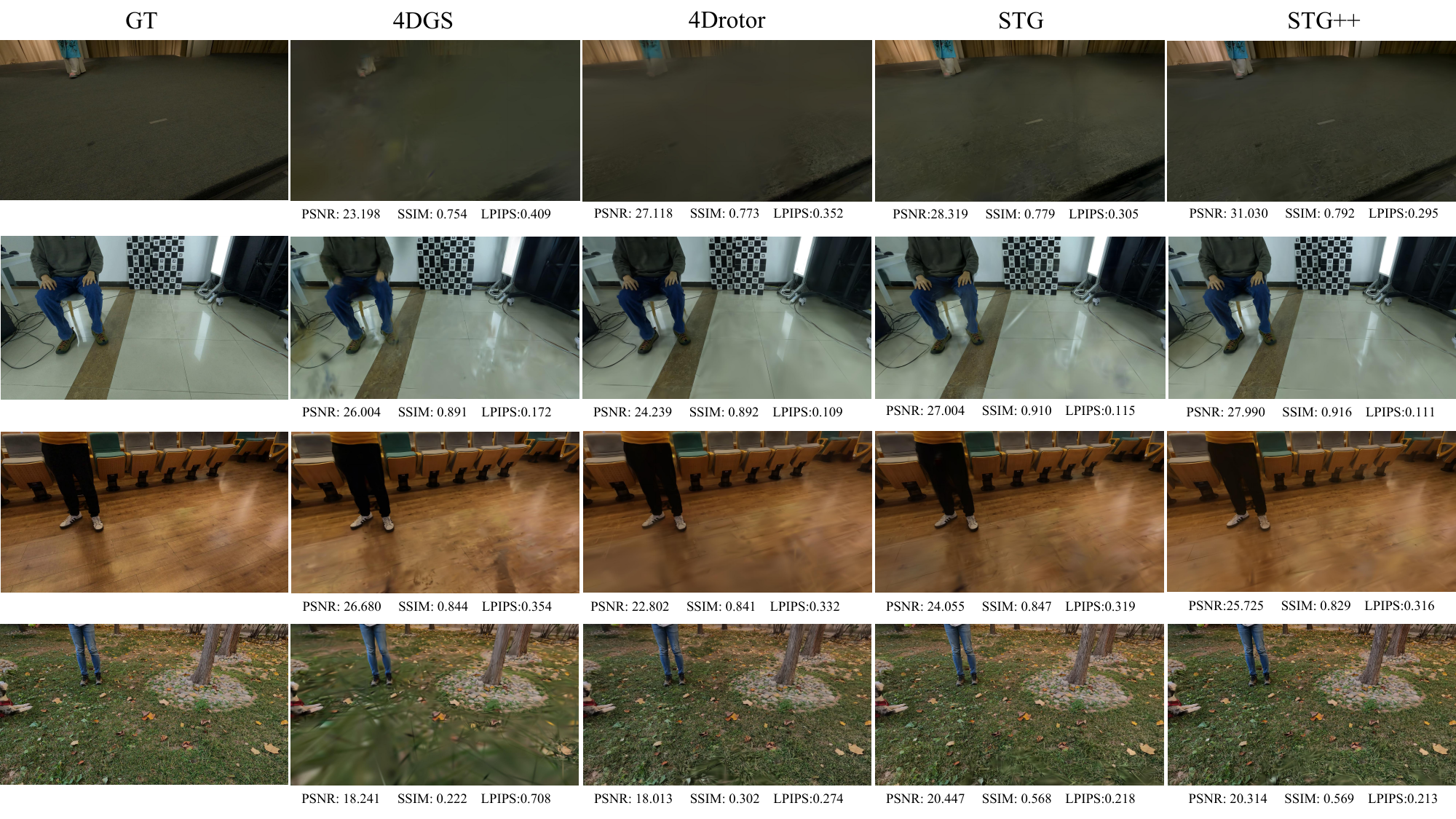}
    \caption{Test view results of three baselines and STG++ on Scene1, Scene2, Scene5, Scene6.}
    \label{fig:testview}
\end{figure*}

\paragraph{Quantitative Results.}
Table~\ref{tab:detail test view results} is an extension of Table~\ref{fig:benchmark}. Table~\ref{tab:trainview} displays the train view performance of baselines and STG++ on each 60-frame segment across different scenes.
\vspace{-10pt}
\paragraph{Qualitative Results.}
Figure~\ref{fig:testview} shows the results of the test view, while Figure~\ref{fig:moretrainview} presents more results of the train view. Here, we also include a comparison with the latest work Ex4DGS~\footnote{Lee J et al. Fully Explicit Dynamic Gaussian Splatting. Advances in Neural Information Processing Systems, 2024, 37: 5384-5409.}, whose code was released shortly before our submission, limiting our ability to investigate thoroughly. Figure~\ref{fig:Ex4DGS_results} presents the Ex4DGS's results from the same viewpoints as Figure 5 showed in the main text. It performs slightly worse in the challenging motion area due to incomplete dynamic and static partitioning. 

\begin{figure*}[htbp]
    \centering
    \begin{subfigure}[b]{\textwidth}
        \includegraphics[width=\textwidth]{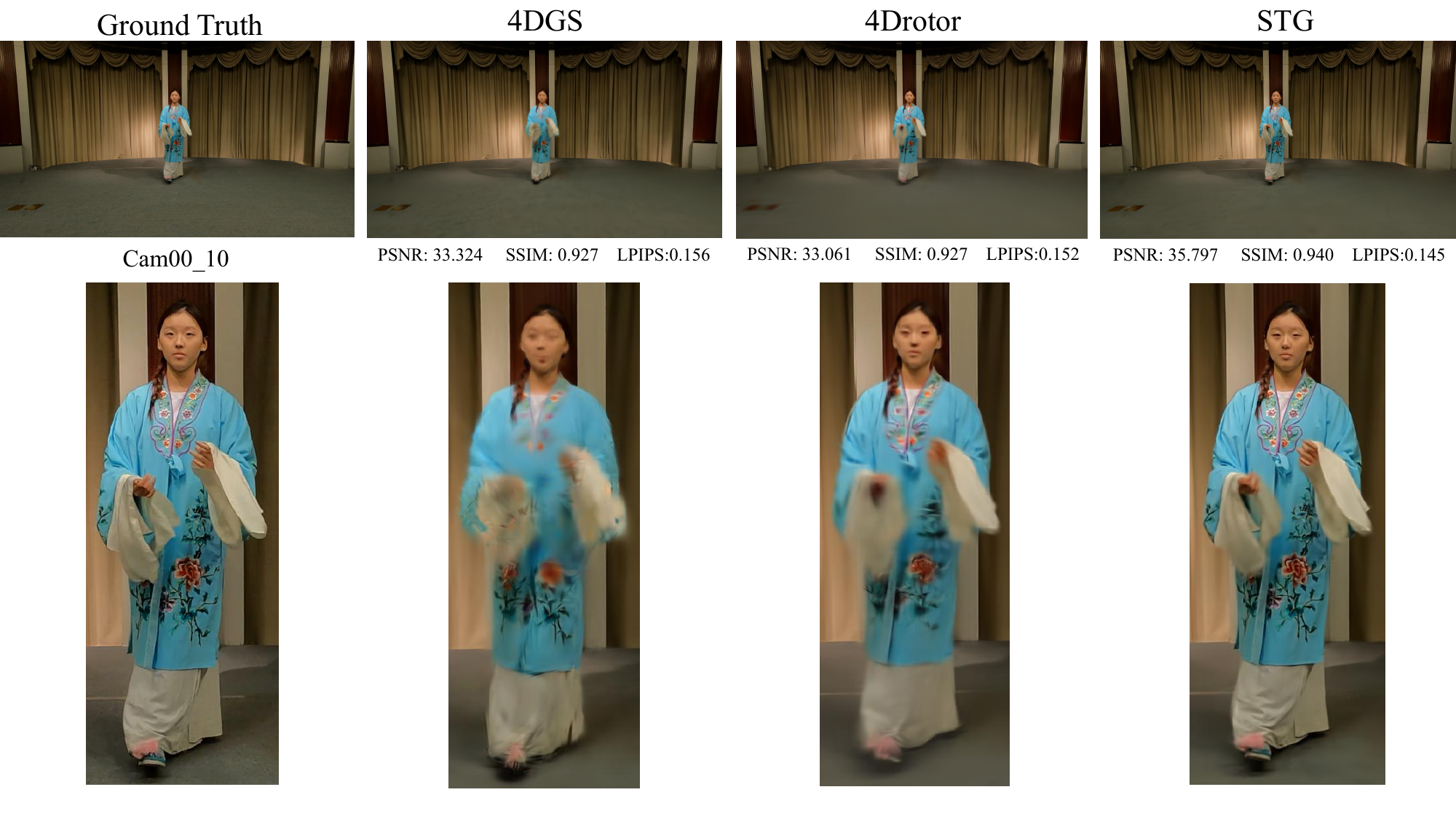}
        \caption{The results of train views for three baselines on \textit{Scene1 Opera\_girl}.}
        \label{fig:subfig1}
    \end{subfigure}
    \vspace{1em}
    \begin{subfigure}[b]{\textwidth}
        \includegraphics[width=\textwidth]{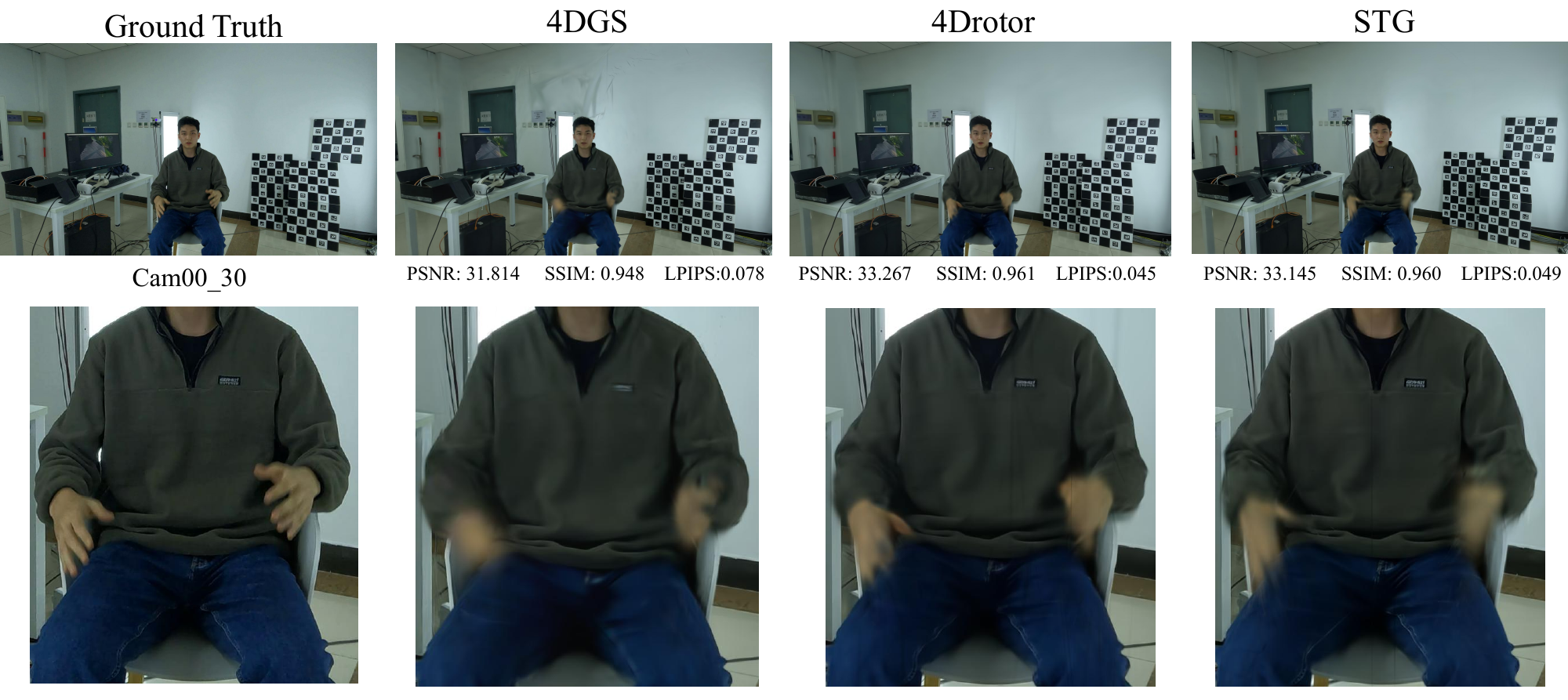}
        \caption{The results of train views for three baselines on \textit{Scene2 Laboratory}.}
        \label{fig:subfig2}
    \end{subfigure}
    \caption{More benchmark results visualization (Part 1).}
    \label{fig:moretrainview}
\end{figure*}

\begin{figure*}[htbp]\ContinuedFloat
    \centering
    \begin{subfigure}[b]{\textwidth}
        \includegraphics[width=\textwidth]{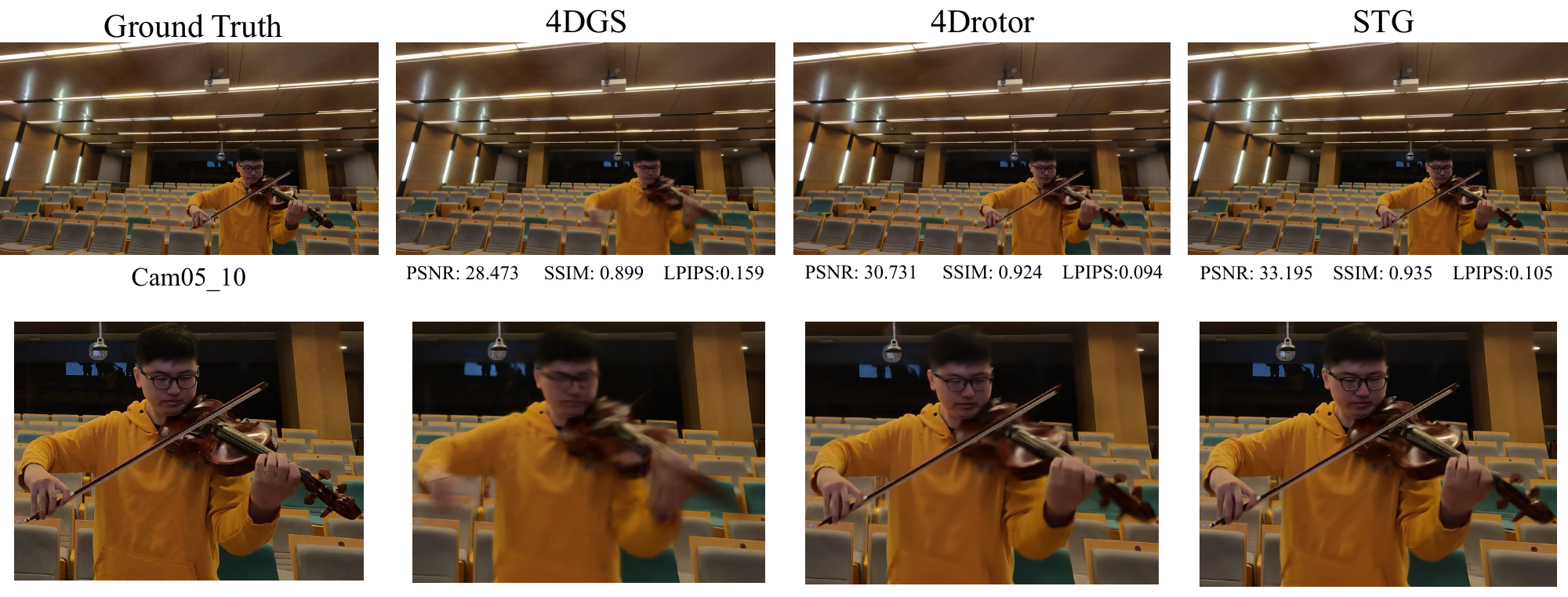}
        \caption{The results of train views for three baselines on \textit{Scene5 Rendition}.}
        \label{fig:subfig3}
    \end{subfigure}
    \vspace{1em}
    \begin{subfigure}[b]{\textwidth}
        \includegraphics[width=\textwidth]{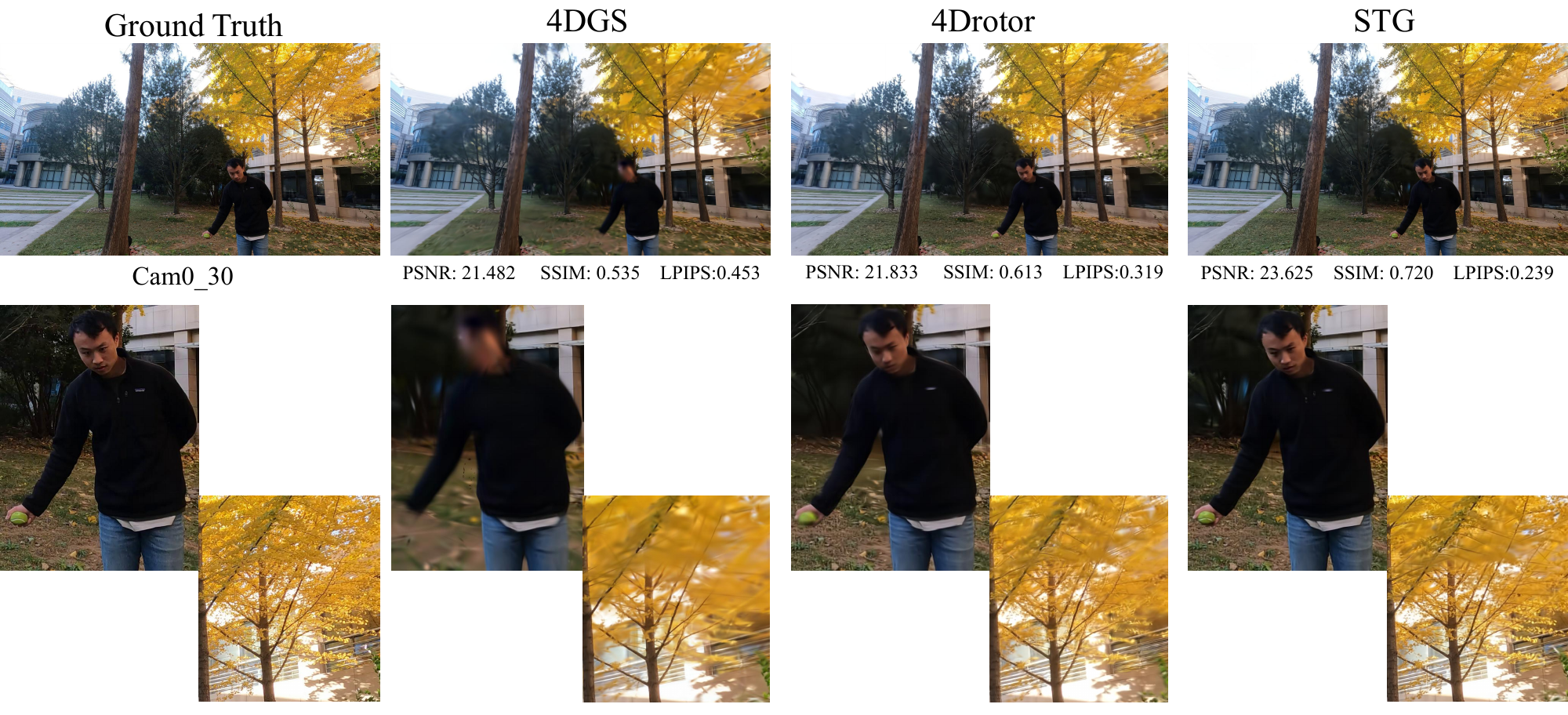}
        \caption{The results of train views for three baselines on \textit{Scene6 Puppy}.}
        \label{fig:subfig4}
    \end{subfigure}
    \caption[]{More benchmark results visualization (Part 2).}
\end{figure*}

\begin{figure}
    \centering
    \includegraphics[width=\linewidth]{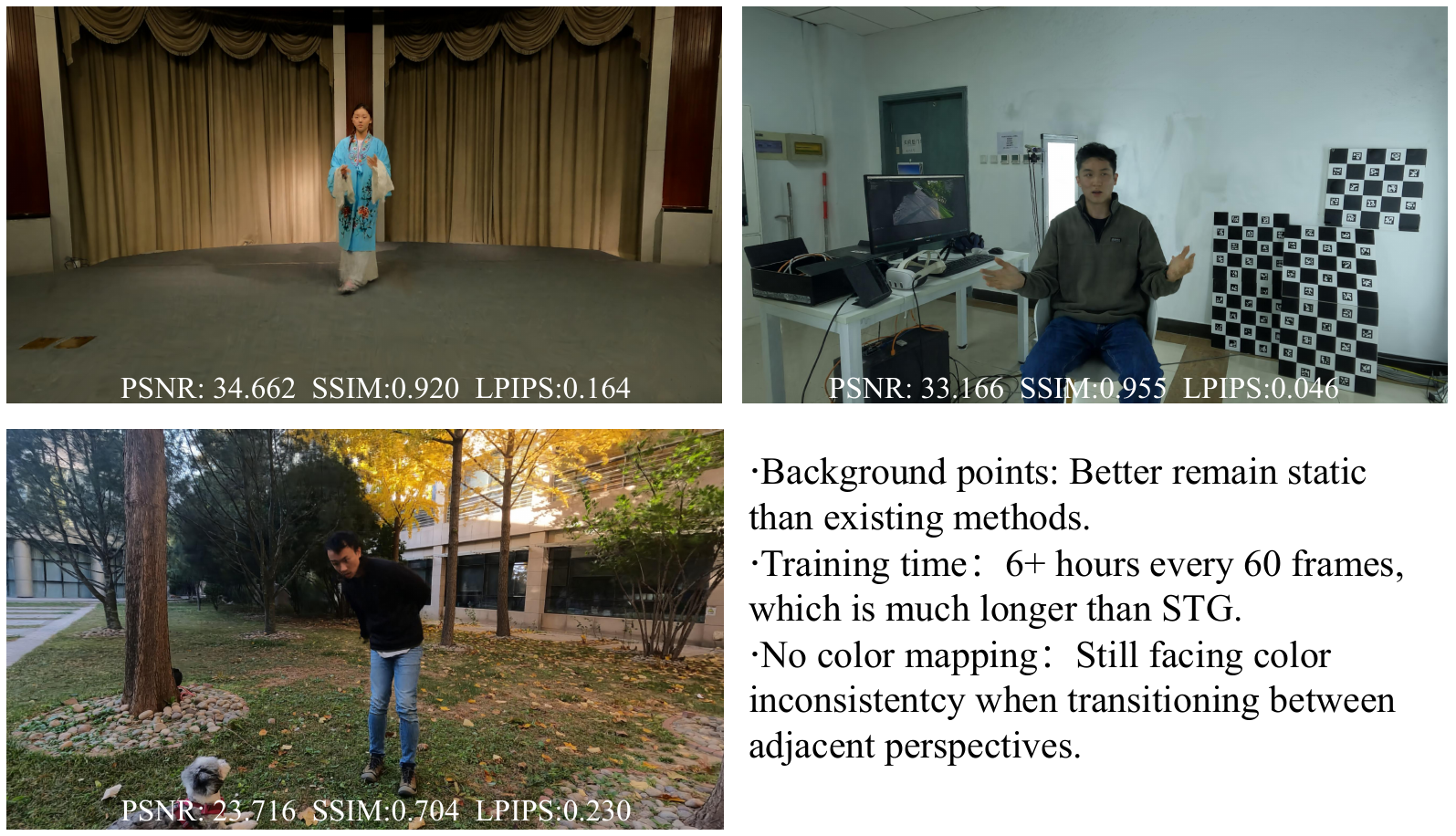}
    \caption{Ex4DGS's performance on the same views as Figure 5 showed in the main text.}
    \label{fig:Ex4DGS_results}
\end{figure}

\subsection{Analysis}
4DGS proposes a two-stage training approach. In the first stage, the algorithm initializes a static scene using the 0th frame and limits the number of final points. It maintains the number and color attributes of the Gaussians while only predicting changes in their positions, rotations, and scaling. This results in minimal model storage, leading to better performance in the static parts of the scene compared to other baselines, with reduced flickering. However, its performance declines significantly in scenes requiring more points for detailed representation, and it cannot address the floaters caused by inconsistent colors in adjacent views. The fitting of larger and faster motions and suddenly-appear/disappear objects is particularly poor.

4Drotor uses dense point clouds as input, which increases memory requirements, especially in large scenes, leading to longer training times and a higher risk of memory overflow. However, by introducing rotors to extend 3D Gaussians to 4D, the authors can directly adapt the density control strategy of the original 3DGS to the t-dimensional space. Consequently, it may perform better in areas with significant motion, such as Figure~\ref{fig:moretrainview} \textit{Scene2 Laboratory} around human hands. 

\section{STG++(Color Mapping) Details}
\label{appendixD}
Although STG is not the smallest in terms of storage among all baselines and cannot directly train a model with 300 frames (requiring splitting into multiple 60-frame segments), it achieves better results under the train-views compared to other baselines. Therefore, we delve deeper into its study, hoping it can serve as the foundational architecture for our initial implementation of immersive volumetric video. However, when viewing in SIBR\_Viewer, we notice two significant drawbacks:

1) In each 60-frame segment, when the viewpoint changes, there is a noticeable flickering of scene points and the presence of floaters, especially when the ground truth of the train-views shows significant color differences due to lighting occlusions and other objective reasons.

2) Besides the color inconsistency during viewpoint changes within each segment, when we modify SIBR\_Viewer to continuously load multiple segments, the transitions between segments become even more abrupt. This is a drawback of segmented training, as the appearance of the Gaussians cannot remain consistent between segments.

Thus, we propose a learnable viewpoint-dependent affine color transformation function $\phi_i(W,T)$ and maintain its values across different segments.
Here, $i$ is the index of the camera, $W$ is a 3×3 transformation matrix, and $T$ is a (1,3) offset vector. Just like the affine transformation, the colors in rendered images $C'_{i}$ are related to the colors in real scenes (SIBR\_Viewer) $C_{i}$  as follows:
\begin{equation}
    C'_{i}  =\mathbf{W} \cdot C_{i}+ \mathbf{T}  
\end{equation}
The loss is calculated between the rendered images $C'_{i}$ and the ground truth as: 
\begin{equation}
    Loss=(1-\lambda_{1} )L_{1} (gt,C'_{i} ) +\lambda_{1}D_{SSIM}(gt,C'_{i}  )
\end{equation}

You can get a more intuitive sense of the improvement from the video in supplementary materials.

\section{Clarifications and Technical Details.}
\label{appendixE}

\subsection{About "Volumetric" Term}
In fact, both academia and industry have yet to provide a clear definition of volumetric video capture methods. Our original intention was to define videos that use 3D reconstruction technologies and provide a 6-DOF experience as volumetric video, which is the future of media. Most existing volumetric videos are constructed from an outside-looking-in manner, often lacking natural backgrounds and lighting, which reduces immersion. Inspired by Google's work~\cite{broxton2020immersive}, we aim to develop videos offering a multi-modal, inside-looking-out 6-DoF experience, and thus call it ``immersive volumetric video".

\subsection{STG++ Limitations}
We introduce viewpoint-based color transformation to address global color inconsistencies caused by varying lighting conditions between cameras in real-world scenes. However, local flickering remains a complex issue due to variations in materials and environmental lighting changes. It is still a significant challenge for the current community, requiring more adaptive and fine-grained processing. We will consider this in future work.

\subsection{More Validation of Sound Field Reconstruction}
It is worth noting that, AV-NeRF~\cite{liang2023av} is the closest existing work to our goal of sound field reconstruction (SFR), which means it also targets sound synthesis in a novel location. However, it focuses solely on sound synthesis and uses professional binaural audio acquisition equipment (which is entirely different from our capture system) to collect sound from various locations in space. As a result, its SFR method cannot directly leverage data collected by our camera rig. We will try to modify and adapt its approach for comparison with our baseline in the future.

\begin{figure}
    \centering
    \includegraphics[width=\linewidth]{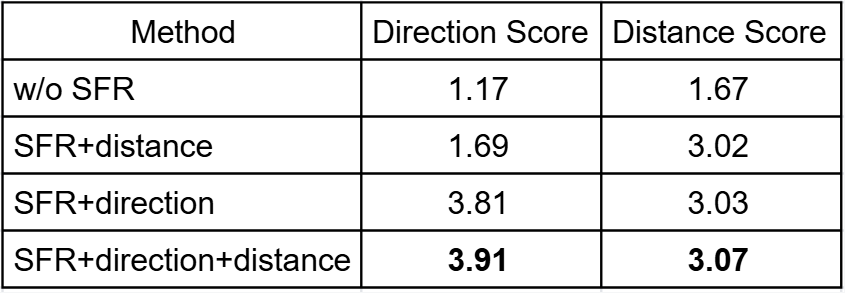}
    \caption{Ablation Studies Illustrate the Effectiveness of Our Proposed Sound Field Reconstruction (SFR) Baseline. A total of 58 participants participated in the user study, rating their sense of direction and distance based on results from various algorithms, using a scale from 1 to 5 (1 being low and 5 being high).}
    \label{fig:sound_field_ablation}
\end{figure}

But to further assess the effectiveness of each module in our proposed SFR method, we have conducted an ablation study based on user feedback, as shown in Figure~\ref{fig:sound_field_ablation}. The average scores for both metrics indicate that incorporating direction and distance modeling in sound field reconstruction significantly enhances participants' immersive video experiences, further demonstrating our SFR algorithm as a practical baseline.

\subsection{Capture Rig Setups and Calibration}
\paragraph{Time-Synchronized.} 
We used GoPro's official QR control app on mobile phone, enabling each camera to scan a dynamically updating QR code for time synchronization.

\paragraph{Noise Reduction and Cart Speed.} 
\begin{figure}
    \centering
    \includegraphics[width=\linewidth]{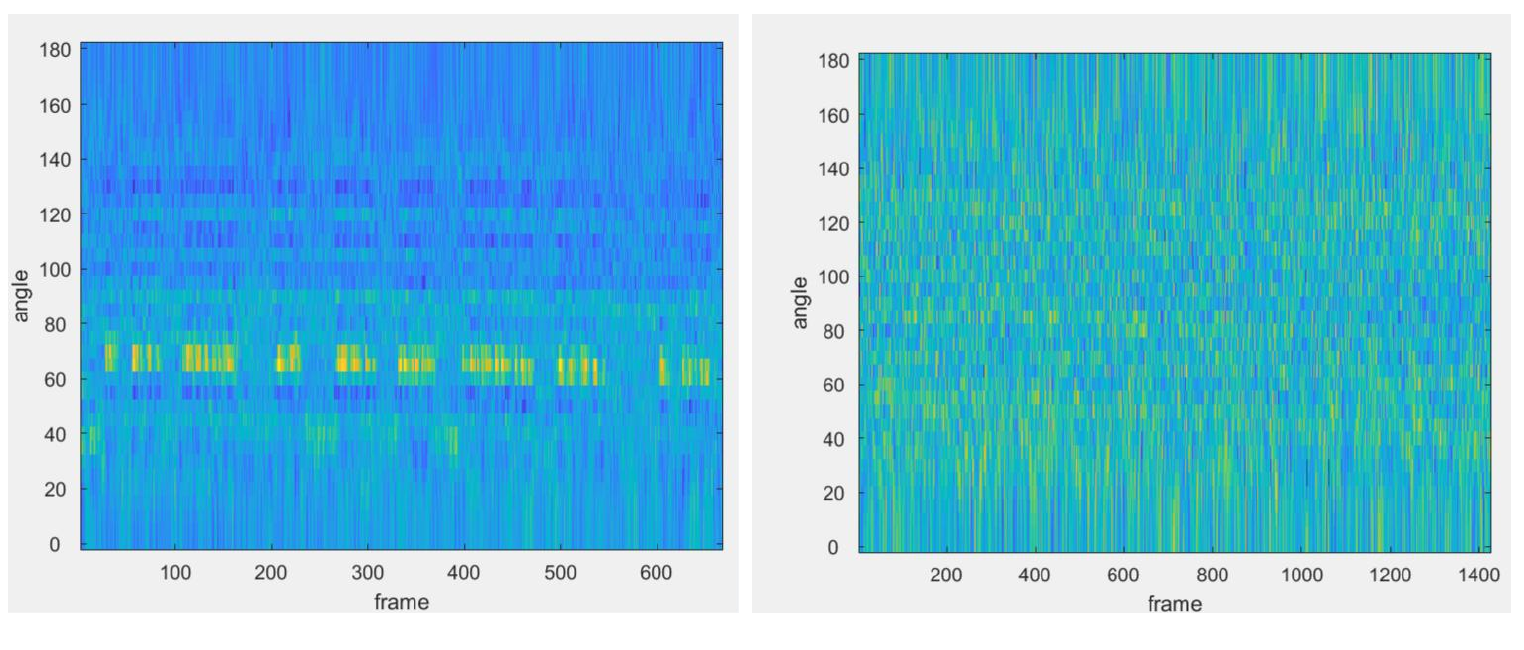}
    \caption{Sound Localization w \& w/o Noise Reduction. The yellow areas in the image represent the highest sound intensity in each frame, clearly indicating the sound source's angle change relative to a specific camera.}
    \label{fig:sound_localization}
\end{figure}
Our cart only generates noise from axle rotation during sharp turns. The description in main text L.250-253 mitigates this noise and greatly reduces the impact of environmental sounds (e.g., wind) on sound quality. As shown in Figure~\ref{fig:sound_localization}, the denoised sound achieves clearer localization (left), while the noisy sound results in more divergent localization (right). This indicates that the sound quality we collected can not only construct multi-modal volumetric videos but also contribute to sound field, inspiring future work on sound localization and reconstruction from multiple sound sources.
Additionally, while the cart experiences slight shaking on uneven terrain, it moves as a rigid body, so its speed is not limited. This prior knowledge can even accelerate our calibration process. The slow speed in this work is primarily due to safety considerations, and we plan to collect faster-moving data in the future.

\paragraph{Calibration.} 
Although we have completed the calibration of the data intended for release, including both fixed-point and mobile shooting, before the submission deadline, it is important to note that, for moving shots, we have tested various open-source algorithms, but none offer an efficient solution for moving multi-view data. Using the original COLMAP takes days to calibrate poses for each frame in long videos. A feasible approach to speed up may refer to \footnote{Bernhard Kerbl et al. A Hierarchical 3D Gaussian Representation for Real-Time Rendering of Very Large Datasets. TOG, 2024.1}. We also look forward to working with colleagues in this community to explore more efficient and accurate calibration solutions using this dataset.

\subsection{Continuously Updated Dataset}
\begin{figure*}
    \centering
    \includegraphics[width=\textwidth]{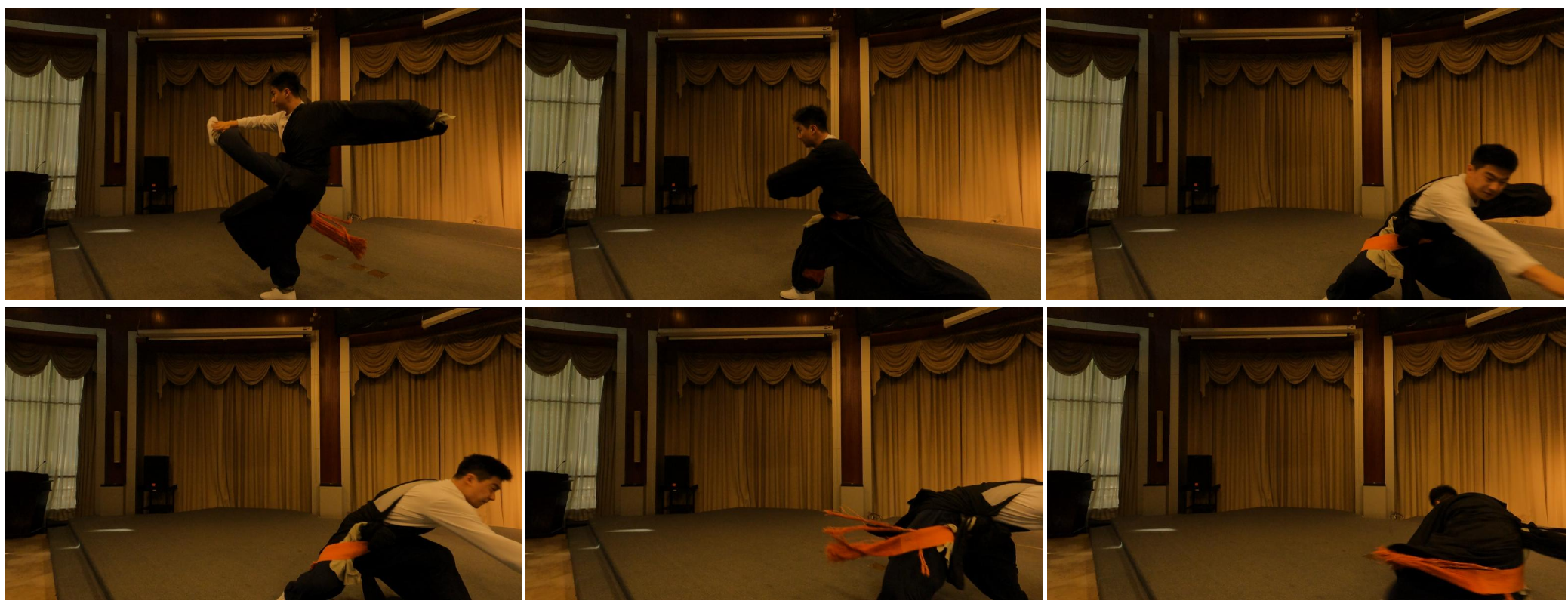}
    \caption{High-Speed Motions Data in Our Dataset ImViD. Scene 1: opera boy spinning kick.}
    \label{fig:fast_motion}
\end{figure*}

Currently, other segments in Scene1 include high-speed motions, as shown in Figure~\ref{fig:fast_motion}. And we will continue to update the dataset, increase its richness to make more contributions to the development of the community.
\begin{table*}[h]
    \centering
    \caption{Performance of three baseline methods and STG++ on the ~\textbf{ImViD} Dataset. All methods selected ~\textbf{cam10} as the test view.}
    \label{tab:detail test view results}
    \resizebox{\textwidth}{!}{%
    \begin{tabular}{lcccccccccccc }
    \toprule
    \multicolumn{1}{c}{}                         & \multicolumn{3}{c}{Scene1 Opera\_girl}                                 & \multicolumn{3}{c}{Scene2 Laboratory}                                  & \multicolumn{3}{c}{Scene5 Violin}                                      & \multicolumn{3}{c}{Scene6 Puppy}                                       \\
    \multicolumn{1}{c}{\multirow{-2}{*}{Method}} & PSNR↑                          & SSIM↑                         & LPIPS↓                        & PSNR↑                          & SSIM↑                         & LPIPS↓                        & PSNR↑                          & SSIM↑                         & LPIPS↓                        & PSNR↑                          & SSIM↑                         & LPIPS↓                        \\ \midrule
    4DGS                                                                 & 23.227                         & 0.753                         & 0.410                         & 25.798                         & 0.889                         & 0.176                         & \cellcolor[HTML]{FE996B}26.586 & 0.842                         & 0.356                         & 18.121                         & 0.222                         & 0.711                         \\
    4DRotor                                                              & 27.263                         & 0.775                         & 0.328                         & \cellcolor[HTML]{FE996B}28.007 & \cellcolor[HTML]{FE996B}0.918 & \cellcolor[HTML]{FE996B}0.098 & 24.083                         & \cellcolor[HTML]{FE996B}0.850 & \cellcolor[HTML]{FE996B}0.296 & 17.916                         & 0.298                         & 0.331                         \\
    STG                                                                  & \cellcolor[HTML]{FFCE93}28.482 & \cellcolor[HTML]{FFCE93}0.786 & \cellcolor[HTML]{FFCE93}0.287 & 26.306                         & 0.910                         & 0.114                         & 23.144                         & \cellcolor[HTML]{FFCE93}0.846 & 0.317                         & \cellcolor[HTML]{FFCE93}20.497 & \cellcolor[HTML]{FFCE93}0.594 & \cellcolor[HTML]{FFCE93}0.211 \\
    STG++                                                                & \cellcolor[HTML]{FE996B}31.240 & \cellcolor[HTML]{FE996B}0.799 & \cellcolor[HTML]{FE996B}0.277 & \cellcolor[HTML]{FFCE93}27.581 & \cellcolor[HTML]{FFCE93}0.916 & \cellcolor[HTML]{FFCE93}0.107 & \cellcolor[HTML]{FFCE93}25.747 & 0.834                         & \cellcolor[HTML]{FFCE93}0.310 & \cellcolor[HTML]{FE996B}20.533 & \cellcolor[HTML]{FE996B}0.598 & \cellcolor[HTML]{FE996B}0.202 \\ \bottomrule
    \end{tabular}%
    }
\end{table*}

\begin{table*}[h]
    \centering
    \renewcommand*{\arraystretch}{1.4}
    \caption{Comparison of average metrics for three baselines and STG++ across four scenes. Due to its smaller model size, 4DGS~\cite{wu20244d} can train 300 frames at once, so there are no segmented results.}
    \label{tab:trainview}
    \resizebox{\textwidth}{!}{%
    \begin{tabular}{llcccccccccccccccccc}
    \toprule
                                        &         & \multicolumn{3}{c}{Frames1-60} & \multicolumn{3}{c}{Frames60-120} & \multicolumn{3}{c}{Frames120-180} & \multicolumn{3}{c}{Frames180-240} & \multicolumn{3}{c}{Frames240-300} & \multicolumn{3}{c}{Avarage}                                                                                               \\
                                        &         & PSNR↑     & SSIM↑   & LPIPS↓   & PSNR↑     & SSIM↑    & LPIPS↓    & PSNR↑      & SSIM↑    & LPIPS↓    & PSNR↑      & SSIM↑    & LPIPS↓    & PSNR↑      & SSIM↑    & LPIPS↓    & PSNR↑                                   & SSIM↑                                  & LPIPS↓                                 \\ \midrule 
                                        & 4DGS    & -         & -       & -        & -         & -        & -         & -          & -        & -         & -          & -        & -         & -          & -        & -         & \textbf{35.005}                         & \cellcolor[HTML]{FE996B}\textbf{0.930} & \textbf{0.156}                         \\
                                        & 4DRotor & 33.502    & 0.912   & 0.142    & 33.735    & 0.913    & 0.137     & 33.749     & 0.913    & 0.137     & 31.460     & 0.893    & 0.155     & 33.718     & 0.913    & 0.139     & \textbf{33.233}                         & \textbf{0.909}                         & \textbf{0.142}                         \\
                                        & STG     & 34.915    & 0.920   & 0.127    & 35.343    & 0.922    & 0.125     & 35.478     & 0.924    & 0.124     & 35.496     & 0.922    & 0.125     & 34.913     & 0.921    & 0.126     & \cellcolor[HTML]{FFCE93}\textbf{35.229} & \textbf{0.922}                         & \cellcolor[HTML]{FFCE93}\textbf{0.125} \\
    \multirow{-4}{*}{Scene1 Opera}      & STG++   & 35.195    & 0.922   & 0.125    & 35.603    & 0.923    & 0.123     & 35.822     & 0.924    & 0.122     & 35.738     & 0.924    & 0.123     & 35.738     & 0.925    & 0.123     & \cellcolor[HTML]{FE996B}\textbf{35.619} & \cellcolor[HTML]{FFCE93}\textbf{0.924} & \cellcolor[HTML]{FE996B}\textbf{0.123} \\ \hline
                                        & 4DGS    & -         & -       & -        & -         & -        & -         & -          & -        & -         & -          & -        & -         & -          & -        & -         & \textbf{32.701}                         & \textbf{0.949}                         & \textbf{0.078}                         \\
                                        & 4DRotor & 36.207    & 0.967   & 0.049    & 36.519    & 0.967    & 0.046     & 34.593     & 0.96     & 0.067     & 36.484     & 0.968    & 0.046     & 36.679     & 0.967    & 0.049     & \cellcolor[HTML]{FE996B}\textbf{36.096} & \cellcolor[HTML]{FE996B}\textbf{0.966} & \cellcolor[HTML]{FE996B}\textbf{0.051} \\
                                        & STG     & 33.405    & 0.949   & 0.077    & 33.257    & 0.949    & 0.078     & 33.641     & 0.951    & 0.077     & 33.140     & 0.948    & 0.081     & 33.298     & 0.950    & 0.078     & \textbf{33.348}                         & \textbf{0.949}                         & \textbf{0.078}                         \\
    \multirow{-4}{*}{Scene2 Laboratory} & STG++   & 33.450    & 0.950   & 0.079    & 33.666    & 0.951    & 0.076     & 33.616     & 0.951    & 0.077     & 33.452     & 0.950    & 0.079     & 33.748     & 0.952    & 0.073     & \cellcolor[HTML]{FFCE93}\textbf{33.586} & \cellcolor[HTML]{FFCE93}\textbf{0.951} & \cellcolor[HTML]{FFCE93}\textbf{0.076} \\ \hline
                                        & 4DGS    & -         & -       & -        & -         & -        & -         & -          & -        & -         & -          & -        & -         & -          & -        & -         & \textbf{33.645}                         & \textbf{0.918}                         & \textbf{0.183}                         \\
                                        & 4DRotor & 33.398    & 0.935   & 0.135    & 33.361    & 0.935    & 0.133     & 33.255     & 0.934    & 0.136     & 33.398     & 0.935    & 0.132     & 32.638     & 0.932    & 0.136     & \textbf{33.210}                         & \cellcolor[HTML]{FE996B}\textbf{0.934} & \textbf{0.134}                         \\
                                        & STG     & 34.508    & 0.930   & 0.158    & 34.029    & 0.929    & 0.161     & 33.900     & 0.928    & 0.165     & 34.178     & 0.929    & 0.163     & 34.222     & 0.929    & 0.161     & \cellcolor[HTML]{FFCE93}\textbf{34.167} & \cellcolor[HTML]{FFCE93}\textbf{0.929} & \cellcolor[HTML]{FFCE93}\textbf{0.162} \\
    \multirow{-4}{*}{Scene5 Rendition}  & STG++   & 34.426    & 0.928   & 0.160    & 34.277    & 0.929    & 0.163     & 34.169     & 0.928    & 0.165     & 34.407     & 0.929    & 0.161     & 34.203     & 0.927    & 0.159     & \cellcolor[HTML]{FE996B}\textbf{34.296} & \textbf{0.928}                         & \cellcolor[HTML]{FE996B}\textbf{0.162} \\ \hline
                                        & 4DGS    & -         & -       & -        & -         & -        & -         & -          & -        & -         & -          & -        & -         & -          & -        & -         & \textbf{21.117}                         & \textbf{0.450}                         & \textbf{0.561}                         \\
                                        & 4DRotor & 21.902    & 0.643   & 0.301    & 21.988    & 0.646    & 0.297     & 21.719     & 0.629    & 0.319     & 21.884     & 0.644    & 0.297     & 21.844     & 0.645    & 0.300     & \textbf{21.867}                         & \textbf{0.641}                         & \textbf{0.303}                         \\
                                        & STG     & 23.307    & 0.714   & 0.247    & 23.381    & 0.717    & 0.243     & 23.387     & 0.718    & 0.241     & 23.331     & 0.717    & 0.245     & 23.413     & 0.718    & 0.241     & \cellcolor[HTML]{FFCE93}\textbf{23.364} & \cellcolor[HTML]{FFCE93}\textbf{0.716} & \cellcolor[HTML]{FFCE93}\textbf{0.243} \\
    \multirow{-4}{*}{Scene6 Puppy}      & STG++   & 23.316    & 0.714   & 0.246    & 23.423    & 0.719    & 0.240     & 23.392     & 0.719    & 0.241     & 23.314     & 0.716    & 0.246     & 23.438     & 0.719    & 0.240     & \cellcolor[HTML]{FE996B}\textbf{23.377} & \cellcolor[HTML]{FE996B}\textbf{0.717} & \cellcolor[HTML]{FE996B}\textbf{0.242} \\ \bottomrule
    \end{tabular}%
    }
\end{table*}

\begin{table*}[h]
    \centering
    \caption{Existing real-world datasets for dynamic novel view synthesis. }
    \label{tab:full datasets comparison}
    \renewcommand*{\arraystretch}{4}
    \resizebox{\textwidth}{!}{%
    \begin{tabular}{m{3.5cm}cccccccccm{7.5cm}}
    \toprule
    \textbf{Datasets}                   & \textbf{No.Scene} & \textbf{Outdoor/Indoor} & \textbf{Cameras}                       & \textbf{Mobility}          & \textbf{Resolution} & \textbf{Angles}                          & \textbf{Duration}  & \textbf{FPS} & \textbf{Multimodality} & \textbf{Content}                                                                                                                                                                                                                       \\
    \midrule
    PanopticSports~\cite{Joo_2017_TPAMI}                      & 65                & Indoor                  & 480 cameras                            & Static                    & 640×480             & 360°                                     & 5mins              & 25           & \XSolidBrush                    & Human-centric actions                                                                                                                                                                                                                  \\
    Technicolor~\cite{sabater2017dataset}                         & 12                & Indoor                  & 16 cameras                             & Static                    & 2048×1088           & Frontal                                  & 2s                 & 30           & \XSolidBrush                     & Has a number of close-ups sequences, captured medium angle scenes and other animated scenes where the movement does not come from a human.                                                                                             \\
    Immersive-Lightfield~\cite{broxton2020immersive}                & 15                & both                    & 46 cameras                             & Static                    & 2560×1920           & Frontal                                  & 10-30s                & 30           & \XSolidBrush                     & Simple and slow motion of human,animals,objects                                                                                                                                                                                        \\
    HyperNeRF~\cite{park2021hypernerf}                           & 17                & Indoor                  & 1 hand-held phone                      & Fixed-point Waving         & 1920×1080           & Frontal                                  & 30-60s             & 30           & \XSolidBrush                     & Waving a mobile phone in front of a moving scene, object-centric                                                                                                                                                                       \\
    Dynamic Scene Datasets (NVIDIA)~\cite{yoon2020novel}      & 8                 & Outdoor                 & \makecell[c]{1 Mobile phone\\/12 cameras}        & \makecell[c]{Fixed-point Waving\\/Static} & 1920×1080           & Frontal                                  & 5s                 & 60           & \XSolidBrush                     & Simple body motions (facial, jump, etc)                                                                                                                                                                                                 \\
    UCSD Dynamic~\cite{lin2021deep}          & 96                & Outdoor                 & 10 cameras                             & Static                    & 1920×1080           & Frontal                                  & 1-2mins            & \textbf{120} & \XSolidBrush                     & Various visual effects and human interactions                                                                                                                                                                                          \\
    ZJU-Mocap~\cite{peng2021neural}                           & 10                & Indoor                  & 21 cameras                             & Static                    & 1024×1024           & 360°                                     & 20s                & 50           & \XSolidBrush                     & Simple body motions (punch, kick, etc.)                                                                                                                                                                                                \\
    Plenoptic Dataset (DyNeRF/Neural 3D)~\cite{li2022neural} & 6                 & Indoor                  & 21 cameras                             & Static                    & 2704×2028           & Frontal                                  & 10-30s                & 30           & \XSolidBrush                     & Contains high specularity, translucency and transparency objects, motions with changing topology, selfcast moving shadows, volumetric effects, various lighting conditions and multiple people moving around in open living room space \\
    D2NeRF~\cite{wu2022d}                              & 10                & Indoor                  & dual-hold phone                        & Fixed-point Waving         & 1920×1080           & Frontal                                  & 5s                 & 30           & \XSolidBrush                     & Contains more challenging scenarios with rapid motion and non-trivial dynamic shadows                                                                                                                                                  \\
    iPhone Datasets~\cite{gao2022monocular}                     & 14                & both                    & \makecell[c]{1 hand-held phone\\/2 cameras }          & \makecell[c]{Fixed-point Waving\\/Static} & 640×480             & Frontal                                  & 8-15s              & 30/60        & \XSolidBrush                     & Featuring non-repetitive motion, from various categories such as generic objects, humans, and pets                                                                                                                                    \\
    Meetroom Datasets~\cite{li2022streaming}                   & 4                 & Indoor                  & 13 cameras                             & Static                     & 1280×720            & Frontal                                  & 10s                & 30           & \XSolidBrush                     & One or three persons have discussion, working, trimming  in a meeting room                                                                                                                                                             \\
    ENeRF-Outdoor~\cite{lin2022efficient}                       & 4                 & Outdoor                 & 18 cameras                             & Static                     & 1920×1080           & Frontal                                  & 20-40s             & 30           & \XSolidBrush                     & Complex human motions                                                                                                                                                                                                                  \\
    Replay~\cite{shapovalov2023replay}                              & 46                & Indoor                  & 12 cameras                             & Static                    & 3840×2160           & 360°                                     & 5mins              & 30           & \checkmark (Audio)             & Dancing, chatting, playing video games, unwrapping presents, playing ping pong                                                                                                                                                         \\
    Campus Datasets~\cite{wang2024masked}                     & 6                 & Outdoor                 & 24 cameras                             & Static                    & 3840×2160           & Frontal                                  & 5-10s              & 30           & \XSolidBrush                     & Includes more realistic observations such as pedestrians, moving cars, and grasses with people playing                                                                                                                                 \\
    MoDGS~\cite{liu2024modgs}                               & 6                 & both                    & 1 cameras                              & Static                    & \textbf{--}         & Frontal                                  & --                 & --           & \XSolidBrush                     & Contains diverse subjects like skating, a dog eating food, YOGA, etc.                                                                                                                                                                  \\
    DiVa-360~\cite{lu2024diva}                               & 53                 & Indoor                    & 53 cameras                              & Static                    & 1280×720         & Frontal                                  & 51s                 & 120           & \checkmark (Audio)                    & For Object-centric tasks. Contains dynamic objects and intricate hand-object interactions.                                                                                                                                                                  \\
    NeRF On-the-go~\cite{ren2024nerf}                      & 12                & both                    & 1 hand-held phone                      & Moveable                   & 4032×3024           & 360°                                  & 5-10s              & 30           & \XSolidBrush                     & Including 10 outdoor and 2 indoor scenes, features a wide range of dynamic objects including pedestrians, cyclists, strollers, toys, cars, robots, and trams), along with diverse occlusion ratios ranging from 5\% to 30\%            \\
    360+X~\cite{chen2024360+}                               & 28                & both                    & \makecell[c]{1 360°cameras and \\1 Spectacles cameras} & Static                    & 5760×2880           & 360°                                     & 10s (2152 sequence) & 30           & \checkmark (Audio)             & Capture in 17 cities across 5 countries.Panoptic perspective to scene understanding with audio                                                                                                                                        \\
    \textbf{ImViD(Ours)}                       & 7                 & \textbf{both}           & \textbf{46 cameras}                    & \textbf{Moveable}          & \textbf{5312×2988}  & \textbf{Frontal and 360°} & \textbf{1-5mins}   & 60           & \textbf{\checkmark (Audio)}    & Seven common indoor and outdoor scenes in daily life, including opera, face-to-face communication, teaching, discussion, music performance, interaction with pets, and playing. Each scene has high-quality synchronized multi-view video and audio with a duration of more than 1 minute, and contains rich elements such as various small objects, glass, and changes in light and shadow                                               \\                  \bottomrule                                                                                                                                                                   
    \end{tabular}%
    }
\end{table*}